\documentclass{datologyai}

\usepackage{url}
\usepackage{svg}
\usepackage{bbding}
\usetikzlibrary{shapes.geometric, positioning, shadows}
\usepackage{lineno}
\usepackage{algorithm}
\usepackage{algpseudocode}

\newif\ifshowvd
\showvdtrue    

\newcommand{\datbench}{%
  \textcolor[HTML]{002ECF}{\textbf{\textsc{DatBench}}}\xspace
}

\newcommand{\datnum}[1]{\textcolor[HTML]{002ECF}{\textbf{#1}}}

\definecolor{baselinegray}{HTML}{2B2B2B}
\def\rolelabelstyle{2}
\newcommand{\curated}{\ifcase\rolelabelstyle curated\or \textit{curated}\or \textcolor[HTML]{002ECF}{\textit{curated}}\fi\xspace}
\newcommand{\Curated}{\ifcase\rolelabelstyle Curated\or \textit{Curated}\or \textcolor[HTML]{002ECF}{\textit{Curated}}\fi\xspace}
\newcommand{\base}{\ifcase\rolelabelstyle baseline\or \textit{baseline}\or \textcolor{baselinegray}{\textit{baseline}}\fi\xspace}
\newcommand{\Base}{\ifcase\rolelabelstyle Baseline\or \textit{Baseline}\or \textcolor{baselinegray}{\textit{Baseline}}\fi\xspace}

\newcommand{\papertitle}{A Prescription for Better VLMs through Data Curation Alone}

\newcommand{\papershortrun}{20/20 Vision Language Models}

\usepackage{soul} 

\projectname{{\fontsize{22.5}{26}\selectfont\textcolor[HTML]{002ECF}{20/20 Vision Language Models:}}}
\projectlogo{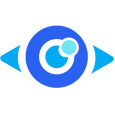}
\title{\papertitle}
\teambyline{DatologyAI Team\footnotemark}
\runningtitle{\papershortrun}
\abstract{%
Data curation has shifted the quality--compute frontier for language-model and contrastive image-text pretraining, but its role for vision-language models (VLMs) is far less established. Here, we ask how far data curation alone can take VLM performance, holding architecture, training recipe, and compute fixed and varying only the training data. Our pipeline, applied to the MAmmoTH-VL single-image subset, lifts performance by \datnum{$+11.7$pp} on average across 20 public VLM benchmarks (spanning grounding, VQA, OCR and documents, captioning, spatial and 3D, counting, charts, math, brand-ID, and multi-image reasoning) and by \datnum{$+11.3$pp} on average across all nine capability axes of \datbench{}~\citep{joshi2026datbench}, our high-fidelity, comprehensive VLM eval suite. At 2B, our \curated model surpasses InternVL3.5-2B by $9.9$pp at \datnum{$\sim 17\times$} less training compute and closes the gap to Qwen3-VL-2B to within $1.8$pp at \datnum{$\sim 87\times$} less compute, all from pretraining alone. Beyond strong gains in accuracy, our curation pipeline delivers four further properties:
(1) \textbf{Reliability}: per-capability standard deviation across training seeds drops by \datnum{$\sim 67\%$} and the curated-vs-baseline lift survives a context-length sweep from 4k to 16k tokens;
(2) \textbf{OOD generalization}: the 9-eval OOD average rises by $+7.2$pp, and multi-image reasoning on BLINK rises by $+3.09$pp despite single-image-only training, with Visual Correspondence gaining $+11.8$pp despite demanding cross-image reasoning;
(3) \textbf{Behavioral gains beyond benchmarks}: across ${\sim}1{,}100$ open-ended queries the \curated 2B is more honest and more specific than the matched-compute \base, and answers more concisely and refuses fewer benign queries than a frontier 2B reference;
(4) \textbf{Pareto-dominance on inference cost}: at every scale tested (1B, 2B, 4B) the \curated model raises accuracy while lowering response FLOPs against the matched-compute \base, and the \curated 4B matches near-frontier accuracy at \datnum{$3.3\times$} lower response FLOPs than Qwen3-VL-4B.
Our pipeline shows that data curation is a high-leverage tool for building better VLMs, reaching near-frontier accuracy at up to \datnum{$\sim 150\times$} less training compute. \looseness=-1
}

\usepackage{tabularx}   
\usepackage{array}      
\usepackage{ragged2e}   

\newcolumntype{L}{>{\RaggedRight\arraybackslash}X}

\begin{document}

\maketitle
\footnotetext{See Contributions and Acknowledgements (\S~\ref{sec:contri}) for full author list.}

\begin{figure}[H]
    \centering
    \captionsetup{skip=2pt}
    \includegraphics[width=0.78\linewidth]{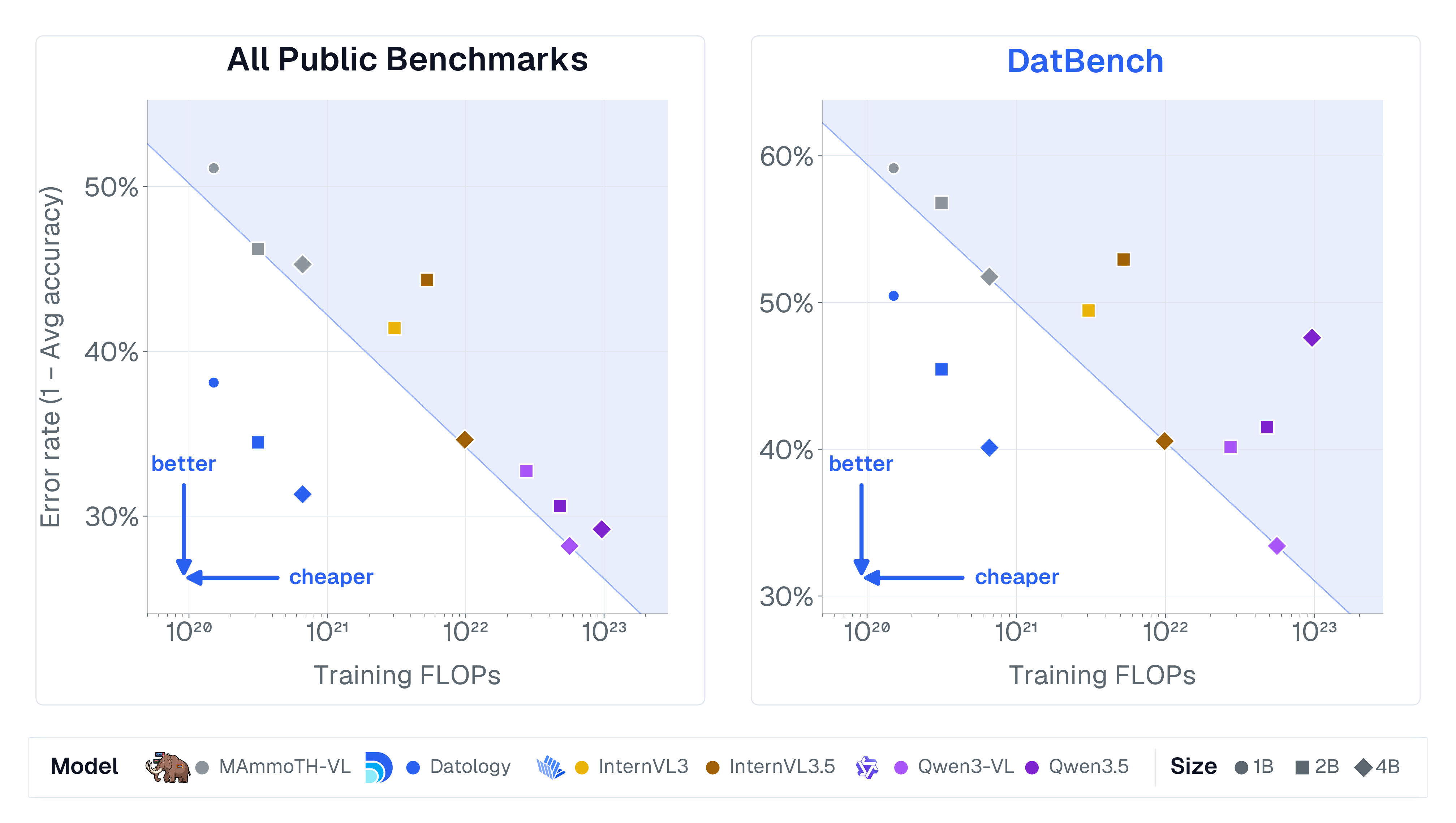}
    \caption{\textbf{Curation matches frontier 2B/4B VLMs at a fraction of the training compute.} Pareto frontier of error rate ($1 - $ avg.\ accuracy) vs.\ training FLOPs on (left) the 20-eval public VLM benchmark suite and (right) \datbench{}. DatologyAI-curated models (blue) at 1B, 2B, and 4B Pareto-dominate the matched-compute MAmmoTH-VL \base (gray) at every scale, and approach Qwen3-VL, InternVL3.5, and Qwen3.5 (extensively post-trained, up to $\sim 150\times$ more compute) on both suites. \looseness=-1}
    \label{fig:hero}
\end{figure}
\clearpage

\section{Introduction}
\label{sec:intro}

In the era of deep learning, curating training data has been a key lever for improvements across modalities. The efficacy of training data curation has been extensively studied in language-model pretraining: curated open corpora (DCLM, FineWeb, RedPajama, Dolma, Nemotron-CC) can shift the quality--compute Pareto frontier by as much as architectural or scale choices \citep{li2024datacomplm, penedo2024fineweb, weber2024redpajama, soldaini2024dolma, su2024nemotroncc, datologyai_text_2024, beyondweb}. Data filtering and mixture design have also become first-class axes in contrastive image-text training  \citep{gadre2023datacomp, xu2023metaclip, fang2023datafilteringnetworks, datologyai_clip_2024}. In vision-language model research, by contrast, the training mixture is typically characterized only at the level of ``X billion image--text pairs'' rather than treated as a primary design variable, with most attention paid to non-data variables such as encoder choice, connector design, resolution handling, visual instruction tuning, and reinforcement learning \citep{liu2023improved, chen2024internvl, wang2024qwen2vl, tong2024cambrian, mckinzie2024mm1, deitke2024molmo, yu2024rlhfv, shen2025vlmr1, liu2025visualrft}. Holding architecture, training recipe, and compute fixed, we measure the impact of pretraining data curation alone on VLM capability.

Prior work on VLM data curation has typically isolated a single axis at a time: multimodal deduplication \citep{slyman2024fairdedup, liu2025ecodatum}, image-text quality filtering \citep{wang2024mlmfilter, wang2025unifilter, chen2024selffilter}, data-mixture design \citep{mckinzie2024mm1, tong2024cambrian, laurencon2024idefics2, deitke2024molmo, shi2024eagle, chen2024far, wang2025openqwen2vl}, recaptioning of paired data \citep{chen2023sharegpt4v, li2024recapdatacomp}, task-agnostic synthetic data \citep{liu2024synthvlm, su2024skvqa, liao2025unicorn}, or task-specific synthetic data \citep{joshi2025mmgenenhancingtaskperformance, yang2025cosyn}. Others release curated corpora without a matched-compute uncurated baseline \citep{guo2024mammothvl}. We instead compose joint image-and-text deduplication, joint image-and-text filtering, target-distribution matching for mixture design, and both task-agnostic and task-specific synthetic data into a single end-to-end pipeline, and ask how far data curation alone can take VLMs.

Concretely, we run our end-to-end data curation pipeline on the single-image subset (10M) of MAmmoTH-VL-12M \citep{guo2024mammothvl} and train models across 1B--4B language backbones (Qwen3 LM \citep{qwen3techreport} with SigLIP2 vision encoder \citep{tschannen2025siglip2}). We compare against a \base trained on the same uncurated MAmmoTH-VL subset under identical training recipe and compute, and against SOTA public models at similar scales: the Qwen3-VL family (2B/4B) \citep{wang2025qwen3vl}, the InternVL3 and InternVL3.5 families (2B/4B) \citep{chen2024internvl}, and Qwen3.5 (2B/4B) \citep{qwen3p5}. These are strong reference points: trained on up to $\sim\!150\times$ more compute and including extensive post-training (instruction tuning, RLHF, RLVR) on top of their pretraining stage.

At 2B, our curation lifts the average across 20 public VLM benchmarks by \datnum{$+11.7$pp}, spanning grounding, VQA, OCR and documents, captioning, spatial and 3D reasoning, counting, charts, math, brand-ID, and multi-image reasoning, and the average across all nine capability axes of \datbench{} by \datnum{$+11.3$pp}. Figure~\ref{fig:hero} previews the cost-quality Pareto: the \curated models Pareto-dominate the matched-compute \base at every scale tested (1B, 2B, 4B), and approach extensively post-trained references at up to \datnum{$\sim 150\times$} less training compute. All results below are post-decontamination, applied to \base and \curated runs alike; the multimodal decontamination methodology is detailed in Appendix~\ref{app:decontamination}.

We organize the paper around five key findings on the impact of data curation for VLMs:
\begin{itemize}
    \item \textbf{Curation shifts the cost-quality Pareto frontier across model scales.} (\S\ref{sec:main-results}) Across the 20-eval public suite, curation improves accuracy at every tested scale: \datnum{$+13.0$pp} at 1B, \datnum{$+11.7$pp} at 2B, and \datnum{$+14.0$pp} at 4B. Across \datbench{}'s nine capability axes, the gains are similarly large, reaching \datnum{$+11.3$pp} at 2B and increasing with scale. Moreover, these gains close the cost-quality gap between pretraining-only curation and extensively post-trained references.

    \item \textbf{Strong OOD generalization.} (\S\ref{sec:main-results}) Our curation lifts the 9-eval OOD average by \datnum{$+7.2$pp} at 2B. Gains extend even to capabilities absent from training: multi-image reasoning on BLINK rises by \datnum{$+3.09$pp} overall, with \texttt{Visual\_Correspondence} gaining \datnum{$+11.8$pp} despite demanding cross-image reasoning.

    \item \textbf{Reliable across seeds and robust across context lengths.} (\S\ref{sec:robustness}) On top of the mean lift, per-capability standard deviation across training seeds drops from $2.47$ to $0.82$pp (\datnum{$\sim 67\%$} reduction), and the advantage survives a context-length sweep from 4k to 16k tokens.

    \item \textbf{Gains carry through to open-ended use beyond benchmarks.} (\S\ref{sec:beyond-benchmarks}) On ${\sim}1{,}100$ open-ended queries spanning OCR, brand and franchise recognition, scene description, and refusal calibration, the \curated 2B is more honest and more specific than the matched-compute \base, and answers more concisely and refuses fewer benign queries than a frontier 2B reference.

    \item \textbf{Pareto-dominant on inference cost.} (\S\ref{sec:inference-efficiency}) \Curated models simultaneously raise accuracy and lower response-FLOPs cost against the matched-compute \base at every scale tested (1B, 2B, 4B). Against extensively post-trained references, the \curated 4B reaches $68.7\%$ average accuracy at \datnum{$3.3\times$} lower response FLOPs than Qwen3-VL-4B, and exceeds Qwen3-VL-2B at \datnum{$1.5\times$} lower response FLOPs.
\end{itemize}

Together, these results show that pretraining data curation alone produces VLM gains that are (1) large, (2) OOD-generalizing, (3) reliable, (4) evident beyond benchmarks, and (5) inference-efficient, at up to \datnum{$\sim 150\times$} less training compute than extensively post-trained references.
\section{Related Work}
\label{sec:related-work}

We organize related work along the three axes most relevant to this work. \emph{VLM modeling} covers the dominant non-data levers vision-language modeling research has emphasized. \emph{VLM evaluation} surveys the benchmark landscape. \emph{Data curation} situates our contribution within the broader curation literature, spanning classical coreset methods, LM and contrastive pretraining, and VLM-specific work.

\paragraph{VLM modeling.} A now-standard recipe couples a pretrained vision encoder with a pretrained language-model backbone via a lightweight projector, trained on interleaved or paired image-text data \citep{liu2023visual, liu2023improved, li2023blip2, alayrac2022flamingo, dai2023instructblip}. Architectural variants differ primarily in how visual tokens are produced and consumed: resampler-style cross-attention \citep{alayrac2022flamingo, laurencon2024idefics2}, linear or MLP projectors over patch tokens \citep{liu2023improved}, dynamic or any-resolution tiling \citep{liu2024llavanext, chen2024internvl, bai2023qwenvl, wang2024qwen2vl, wang2025qwen3vl}, and mixture-of-encoder designs \citep{tong2024cambrian, shi2024eagle, mckinzie2024mm1}. Recent open-weight families in the 2B--8B class (the Qwen-VL line, InternVL, Idefics, Molmo, MM1, PaliGemma, Cambrian, NVLM) \citep{wang2024qwen2vl, wang2025qwen3vl, chen2024internvl, chen2024far, laurencon2024idefics2, laurencon2024idefics3, deitke2024molmo, mckinzie2024mm1, beyer2024paligemma, tong2024cambrian, dai2024nvlm} center most of their work on encoder choice, connector design, resolution handling, and post-training, with a parallel line targeting capabilities downstream of pretraining via preference optimization and reinforcement learning across hallucination, visual reasoning, grounding, counting, and spatial tasks \citep{yu2024rlhfv, sun2023aligning, zhou2024povid, li2024silkie, huang2025visionr1, shen2025vlmr1, meng2025mmeureka, liu2025visualrft}. Several of the open-weight families do include mixture or data-inclusion ablations as pre-work \citep{mckinzie2024mm1, tong2024cambrian, deitke2024molmo, laurencon2024idefics2}, but these stay inside a single model recipe and reduce the data question to whether one mixture improves an aggregate benchmark score. Our focus is the complementary question of isolating pretraining data as a design variable while holding architecture, recipe, and compute fixed.

\paragraph{VLM evaluation.} Existing benchmarks cover knowledge and reasoning \citep{yue2024mmmu, chen2024mmstar, liu2023mmbench, fu2023mme, li2023seedbench, yu2023mmvet}, perception and multi-image reasoning \citep{fu2024blink, tong2024cambrian, xai2024realworldqa}, optical character recognition (OCR) and document understanding \citep{liu2024ocrbench, mathew2021docvqa}, hallucination \citep{li2023pope}, counting \citep{paiss2023countbench, acharya2019tallyqa}, and grounding \citep{kazemzadeh2014refcoco, yu2016refcocog}. Parallel work has documented saturation, prompt sensitivity, and contamination in these suites \citep{chen2024mmstar, tong2024eyeswideshut, adiga-etal-2025-attention}, motivating more targeted diagnostic benchmarks. \datbench{} \citep{joshi2026datbench} builds on this line and is the evaluation instrument we use throughout the paper.

\paragraph{Data curation.} In contrastive and LM pretraining, data curation has become a first-class component of the recipe and can deliver large gains at fixed compute. Classical coreset and subset-selection methods select training examples to preserve learning dynamics or gradients \citep{killamsetty2021glister, killamsetty2021retrieve, adiga-etal-2024-designing, pmlr-v202-joshi23b, pmlr-v238-joshi24a, joshi2025datasetdistillationknowledgedistillation}; deduplication and quality filtering scale these ideas to web data \citep{abbas2023semdedupdataefficientlearningwebscale, tirumala2023d4, lee2022deduplicating, merrick2026luxical}. Curated open corpora such as DCLM, FineWeb, RedPajama, Dolma, Nemotron-CC, and \"UberWeb have made curation central to LM pretraining \citep{li2024datacomplm, penedo2024fineweb, weber2024redpajama, soldaini2024dolma, su2024nemotroncc, carranza2026uberweb, datologyai_text_2024, beyondweb}, with parallel progress in contrastive image-text training \citep{gadre2023datacomp, xu2023metaclip, fang2023datafilteringnetworks, datologyai_clip_2024} and extensions to objectives beyond raw quality such as safety pretraining \citep{maini2025safetypretraininggenerationsafe}. Complementary evidence suggests that much of the leverage for downstream quality lies in the pretraining mixture itself \citep{baek2026finetunersfallacy}. 

In VLMs, by contrast, prior curation work has mostly studied one axis at a time: multimodal deduplication \citep{slyman2024fairdedup, liu2025ecodatum}, image-text quality filtering \citep{wang2024mlmfilter, wang2025unifilter, chen2024selffilter}, recaptioning of paired data \citep{chen2023sharegpt4v, li2024recapdatacomp, lai2025veclip}, task-agnostic and task-specific synthetic data \citep{liu2024synthvlm, su2024skvqa, liao2025unicorn, joshi2025mmgenenhancingtaskperformance, yang2025cosyn}, mixture design within a single-model recipe \citep{mckinzie2024mm1, tong2024cambrian, laurencon2024idefics2, deitke2024molmo, wang2025openqwen2vl}, or releases of curated corpora without a matched-compute reference \citep{guo2024mammothvl, finevision2025}. Recent work has begun composing two of these axes \citep{liu2025ecodatum, wang2025openqwen2vl}, but the compositions remain partial and the dominant evaluation pattern reduces the data question to a single benchmark average. Our contribution is to treat pretraining data as the primary variable of interest under matched architecture, recipe, and compute, and to study how curation affects not just benchmark averages but also robustness, generalization beyond benchmarks, and inference-time efficiency.
\section{Background}
\label{sec:background}
VLM development is commonly partitioned into four phases \citep{liu2023visual, liu2023improved, bai2023qwenvl, wang2024qwen2vl, wang2025qwen3vl, chen2024internvl, tong2024cambrian, mckinzie2024mm1, deitke2024molmo, shi2024eagle, laurencon2024idefics2, laurencon2024idefics3}: \emph{VLM pretraining}, which takes a pretrained language model (LM) and a pretrained vision encoder and adapts them into a single joint vision-language model via large-scale image-text training; \emph{supervised / visual instruction fine-tuning (SFT)}, which adapts the pretrained VLM to specific task formats; \emph{preference optimization}, including reinforcement learning from human feedback (RLHF) and direct preference optimization (DPO), which aligns model outputs to human preference data; and \emph{reinforcement learning with verifiable rewards (RLVR)}, which optimizes against verifiable reward signals such as math correctness or code execution. VLM pretraining is the focal phase for this paper: it establishes cross-modal alignment and forms the base on which all downstream stages build, and the pretraining mixture remains where much of the downstream leverage lies \citep{baek2026finetunersfallacy}. Every intervention in this paper is confined to pretraining data curation: we run no SFT, no preference optimization, and no RLVR. This makes the comparison against Qwen3-VL and InternVL3.5 a strong one: those public models include all three post-training stages on top of their pretraining, while ours does not.

\subsection{Setup}

\paragraph{Model.} The vision encoder is SigLIP2-SO400M at $384{\times}384$ patch-14 resolution \citep{tschannen2025siglip2}, and the language backbone is drawn from the Qwen3 family \citep{qwen3techreport}. We train models at three scales, \textit{1B}, \textit{2B}, and \textit{4B}, each pairing a Qwen3 language-model backbone with SigLIP2-SO400M. High-resolution images are consumed at native resolution via dynamic tiling in the style of InternVL \citep{chen2024internvl}: each image is split into an aspect-ratio-aware grid of $384{\times}384$ tiles (up to $12$ per image) plus a thumbnail, so the total visual-token budget scales with resolution. SigLIP2 encodes each tile into $729$ tokens; a $1{\times}3$ pixel shuffle then compresses each tile to $243$ tokens while expanding feature width from $1{,}152$ to $3{,}456$ channels, and a two-layer MLP projector maps visual features into the language backbone.

\paragraph{Training recipe.} All models are trained for $25$B tokens in a single stage (projector and LM are trained jointly) at a default context length of $4{,}096$ tokens, with global batch size $512$ in $12$k steps; sequences are packed to minimize padding. The robustness experiments in \S\ref{sec:robustness} sweep the context length to 8k and 16k while keeping the token budget constant. Optimizer, schedule, warmup, weight decay, precision, batch size, and hardware are standard and reported in full in Appendix~\ref{app:hparams}.

\subsection{Data}

\paragraph{\Base: MAmmoTH-VL (single-image subset).} Our \base corpus is the single-image subset of MAmmoTH-VL-12M \citep{guo2024mammothvl}, roughly $10$M samples after filtering out multi-image instances. MAmmoTH-VL itself aggregates open-source instruction data across natural-image VQA, OCR and document understanding, charts and tables, math and science, captioning, and referring expressions, and supplements them with a re-generation pass in which open VLMs and LMs rewrite annotations into longer, chain-of-thought-style responses; it is broad in coverage and representative of current open VLM training mixtures. We restrict the scope of this study to single-image samples for training, while showing that the resulting curation improvements generalize to multi-image evaluations as well.

\paragraph{DatologyAI-curated mixture.} The \curated mixture is produced by applying the DatologyAI curation pipeline on top of the single-image MAmmoTH-VL subset, matched to the \base in \emph{total training tokens}; we summarize the VLM-specific components below.

\emph{Deduplication.} MAmmoTH-VL aggregates many open-source datasets, so both images and text are heavily duplicated across sources. We apply multimodal deduplication to reduce redundant examples and improve control over the resulting training mixture.

\emph{Mixture design and distribution matching.} We analyze the source corpus along several multimodal axes and adjust the resulting mixture to improve coverage, balance, and downstream utility. This includes distribution-level interventions that account for both image and text properties, while preserving broad coverage of the original corpus.

\emph{Filtering.} We apply multimodal quality filters that operate jointly on image and text signals to remove low-signal, malformed, or off-distribution samples. These filters are tuned to retain diverse and rare-but-useful examples rather than simply maximizing average per-sample quality in isolation.

\emph{Synthetic data.} Two families of synthetic data complement the filtered, remixed corpus: a \emph{task-agnostic} pipeline that broadens coverage of the source corpus, and \emph{task-specific} generation pipelines for selected capability families.

\subsection{Evaluation}

We report on two complementary evaluations: \datbench{} \citep{joshi2026datbench}, the high-fidelity VLM eval suite presented in prior work, and 20 public VLM benchmarks.

\paragraph{\datbench{}.} \datbench{} draws from over 30 source benchmarks, transforming questions into generative formats where possible, removing samples solvable without the image, and filtering mislabeled or ambiguous examples, to yield roughly $5{,}000$ samples per capability. Coverage spans nine capability axes (grounding, chart, scene, table, spatial, math, counting, document, and general); per-capability scores are averaged into a single summary metric.

\paragraph{IID vs.\ OOD.} IID and OOD are defined with respect to the curation pipeline, not with respect to whether a benchmark is public or to its high-level domain. IID benchmarks correspond to task formats that explicitly informed task-specific synthetic generation or mixture design; OOD benchmarks were not used to shape those curation decisions, even when they probe similar capabilities. BLINK is OOD on a stronger axis: it requires reasoning over multiple images, and our curation is single-image only.

\paragraph{Public benchmark suite.} We report on 20 public VLM benchmarks, split into 11 IID and 9 OOD evaluations.

\textbf{IID public benchmarks} cover referring and grounding, general VQA, OCR and document understanding, counting, chart and diagram reasoning, and math: RefCOCO, RefCOCOG, RefCOCO+, PixMo Points, RealWorldQA, TextVQA, DocVQA, CountBench, AI2D, ChartQA, and MathVista.

\textbf{OOD public benchmarks} cover general VQA, OCR and document understanding, captioning, spatial and 3D reasoning, domain-specific recognition, and multi-image reasoning: MMBench, OCRBench, DetailCaps, CAPability, CVBench-2D, CVBench-3D, 3DSRBench, ecommerce brand-ID, and BLINK.

All training data is decontaminated against the full evaluation suite using the procedure described in Appendix~\ref{app:decontamination}.
\section{Data Curation Improves VLMs Across Benchmarks, Domains, and Scales}
\label{sec:main-results}

We report the main result through five complementary views. First, at 2B, curation delivers a large average lift on the 11 IID benchmarks of the public suite. Second, the lift transfers out-of-distribution across nine OOD benchmarks, including domain shift to ecommerce brand-ID and structural shift in BLINK multi-image reasoning despite single-image training. Third, \datbench{} shows the gains span all nine capability axes rather than a narrow subset. Fourth, a grounding deep-dive checks that the headline $+57.1$pp gain is a bona fide capability shift rather than a scoring artifact. Finally, scaling runs at 1B, 2B, and 4B show the curation effect persists across model size. Full per-benchmark, per-seed numbers are in Appendix~\ref{app:results}; shared inference settings are in Appendix~\ref{app:inference}.

\begin{figure}[h]
    \centering
    \includegraphics[width=\linewidth]{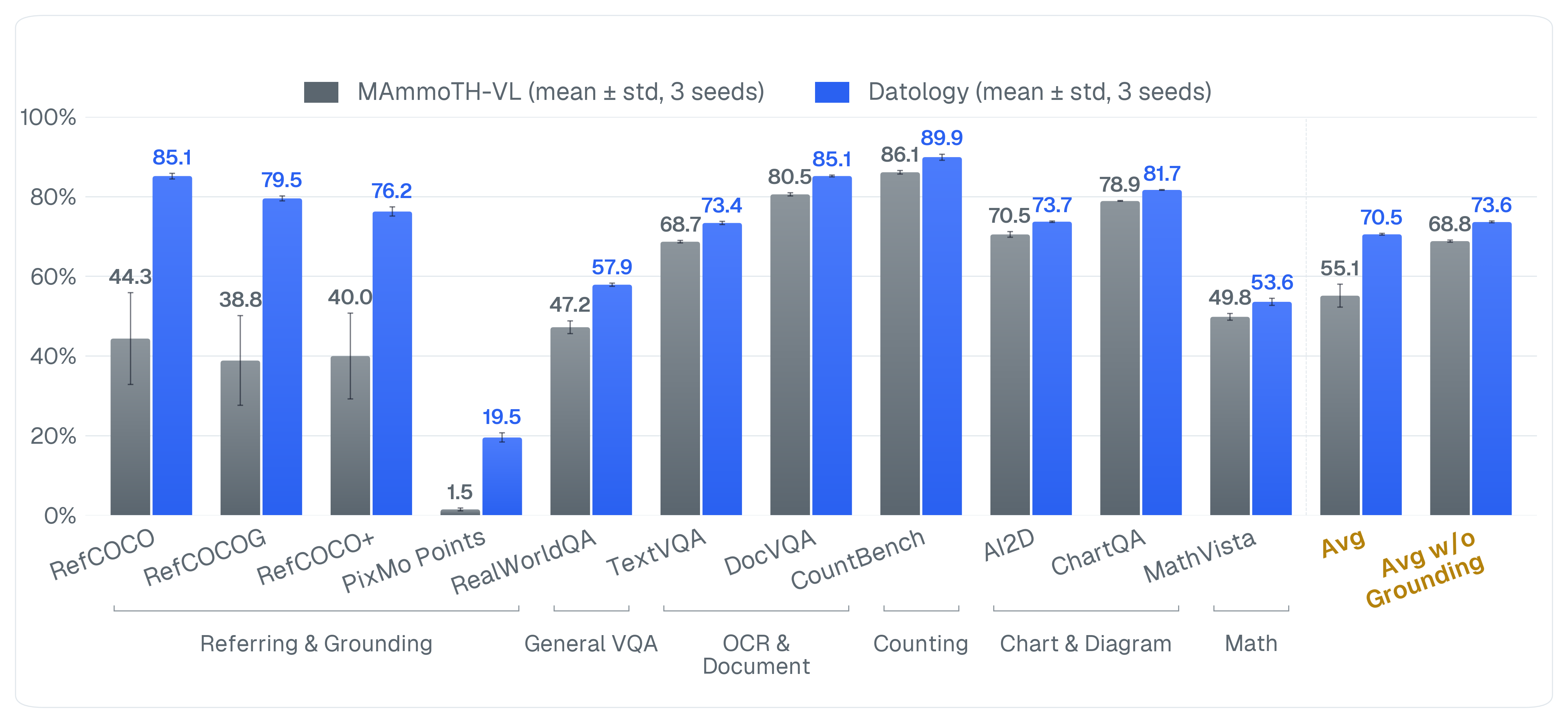}
    \caption{\textbf{Curation lifts the 11 IID public-suite benchmarks by $+15.4$pp on average.} \Curated vs.\ \base on the 11 IID public benchmarks (the 20-eval public suite minus the 9 OOD benchmarks shown in Figure~\ref{fig:ood-capability-bars}) spanning referring \& grounding, general VQA, OCR \& document understanding, counting, chart \& diagram reasoning, and math. Deltas annotated above each pair; the rightmost columns are the 11-IID average ($55.1 \rightarrow 70.5$, $+15.4$pp) and the same average excluding the four referring \& grounding evals ($68.8 \rightarrow 73.6$, $+4.8$pp).}
    \label{fig:general-capability-bars}
\end{figure}

\paragraph{Lift across the 11 IID public-suite benchmarks.}
Figure~\ref{fig:general-capability-bars} reports the curated-vs-baseline comparison on the 11 IID benchmarks in the public suite, spanning referring \& grounding, general VQA, OCR \& document understanding, counting, chart \& diagram reasoning, and math. The 11-IID average rises from $55.1$ to $70.5$ ($\mathbf{+15.4}$pp). Excluding the four referring \& grounding evals, the 7-eval IID average still lifts from $68.8$ to $73.6$ ($\mathbf{+4.8}$pp): the headline does not rest on grounding alone, which we examine separately in \S\ref{subsec:grounding-deep-dive}.

\begin{figure}[t]
    \centering
    \includegraphics[width=\linewidth]{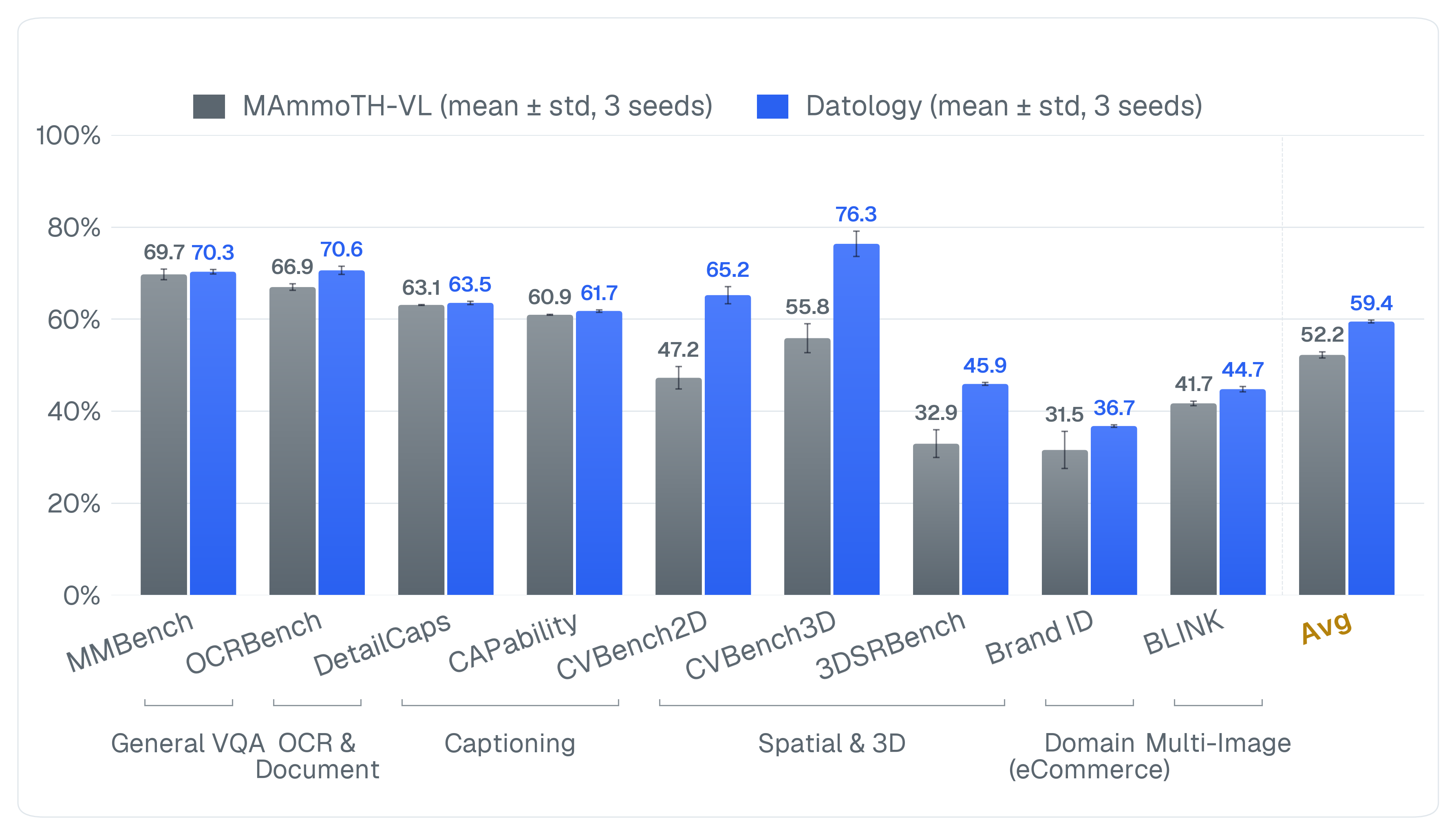}
    \caption{\textbf{Curation gains transfer out-of-distribution.} \Curated vs.\ MAmmoTH-VL \base at 2B on 9 out-of-distribution (OOD) public benchmarks spanning general VQA, OCR \& document, captioning, spatial \& 3D reasoning, domain-specific ecommerce brand-ID, and multi-image reasoning (BLINK). Each bar is averaged over three seeds; error bars show $\pm 1$ standard deviation. The rightmost column is the OOD 9-eval average ($52.2 \rightarrow 59.4$, $+7.2$pp). Every OOD benchmark improves under curation, despite none of these benchmarks being used to inform task-specific synthesis or mixture decisions.}
    \label{fig:ood-capability-bars}
\end{figure}

\paragraph{Out-of-distribution generalization.}
The lift transfers beyond the benchmarks the curation pipeline was tuned on. Across 9 OOD benchmarks (general VQA and OCR/document understanding via MMBench and OCRBench, captioning via DetailCaps and CAPability, spatial \& 3D reasoning via CVBench-2D, CVBench-3D, and 3DSRBench, domain-specific ecommerce brand-ID, and multi-image reasoning via BLINK), the \curated model lifts the average by $+7.2$pp ($52.2 \rightarrow 59.4$), and \emph{every} OOD benchmark improves (Figure~\ref{fig:ood-capability-bars}). We single out two cases that sharpen the OOD claim: ecommerce brand-ID, a domain shift, and BLINK, a structural shift from single-image training to multi-image reasoning.

\begin{figure}[t]
    \centering
    \includegraphics[width=\linewidth]{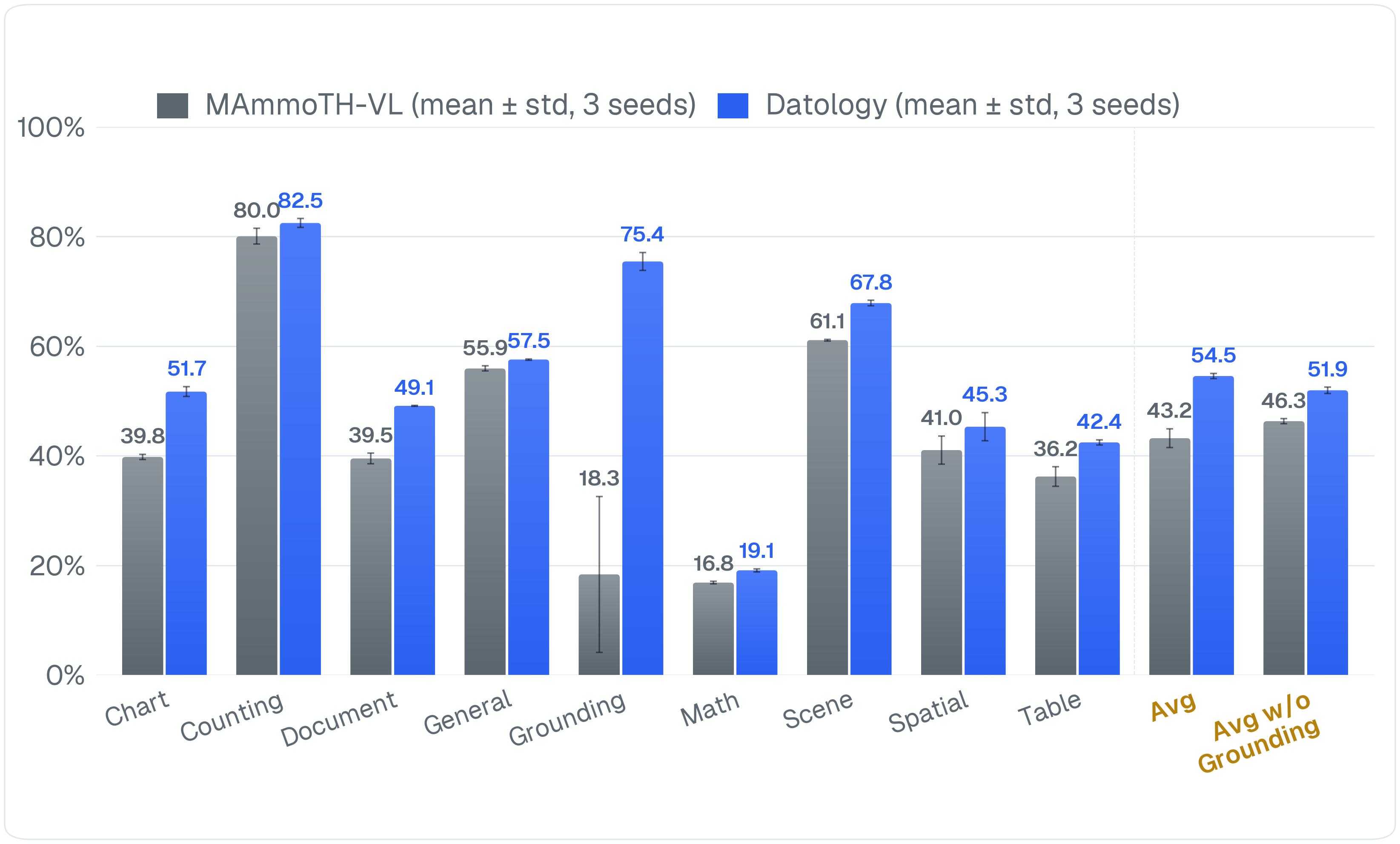}
    \caption{\textbf{Curation lifts every \datbench{} capability, not a favored few.} Per-capability \datbench{} scores at 2B, MAmmoTH-VL \base (gray) vs.\ DatologyAI-curated (blue), each bar averaged over three seeds; error bars show $\pm 1$ standard deviation. The rightmost columns are the 9-capability average ($43.2 \rightarrow 54.5$, $+11.3$pp) and the same average excluding Grounding ($46.3 \rightarrow 51.9$, $+5.6$pp). Grounding carries the largest single gain ($+57.1$pp), and all other capability axes also improve: Chart Understanding ($+11.9$pp), Document Understanding ($+9.6$pp), Scene OCR ($+6.7$pp), Diagrams \& Tables ($+6.2$pp), Spatial Reasoning ($+4.3$pp), Counting ($+2.5$pp), Math \& Logic ($+2.3$pp), and General ($+1.6$pp).}
    \label{fig:main-capability-bars}
\end{figure}

\textbf{Ecommerce brand-ID.} The ecommerce brand-ID benchmark, derived from the Shopify product-catalogue dataset \citep{shopify2024productcatalogue}, is the cleanest single-domain OOD test in our setup: it targets hyper-specific brand identification from product images, a domain the general-purpose curation pipeline was not designed to address. The \curated model improves from $31.5$ to $36.7$ ($+5.2$pp), evidence that the curation effect transfers well beyond the benchmarks the pipeline was tuned on.

\textbf{Single-image to multi-image transfer (BLINK).} BLINK requires evidence from more than one image, spanning categories such as relative depth and visual correspondence as well as counting and object localization; none of these multi-image comparisons appear in our single-image curation. Aggregate accuracy rises from $41.66 \pm 0.59\%$ (\Base) to $44.75 \pm 0.67\%$ (\Curated), a gain of $+3.09$pp, and every \Curated seed exceeds every \Base seed. At the category level, the largest gains are in \texttt{Relative\_Depth} ($59.1\% \rightarrow 77.4\%$, $+18.3$pp) and \texttt{Visual\_Correspondence} ($27.1\% \rightarrow 39.0\%$, $+11.8$pp); the latter directly demands cross-image reasoning, which the single-image pipeline was never trained for.

\begin{takeaway}
\textbf{Data curation generalizes out-of-distribution.} Single-image curation lifts the 9-eval OOD average by \datnum{$+7.2$pp}, improves every OOD benchmark, and transfers to BLINK multi-image reasoning despite no multi-image data appearing in training.
\end{takeaway}

\paragraph{Improvements across all nine \datbench{} capability axes.}
Figure~\ref{fig:main-capability-bars} reports per-capability \datbench{} scores at 2B for the MAmmoTH-VL-12M single-image \base and the DatologyAI-curated mixture, averaged over three seeds. Curation raises the 9-capability average from $43.2$ to $54.5$ ($\mathbf{+11.3}$pp), with improvements on every capability axis. The largest gain is on \texttt{Grounding} ($+57.1$pp), followed by \texttt{Chart Understanding} ($+11.9$pp), \texttt{Document Understanding} ($+9.6$pp), \texttt{Scene OCR} ($+6.7$pp), \texttt{Diagrams \& Tables} ($+6.2$pp), \texttt{Spatial Reasoning} ($+4.3$pp), \texttt{Counting} ($+2.5$pp), \texttt{Math \& Logic} ($+2.3$pp), and \texttt{General} ($+1.6$pp). Excluding Grounding, the remaining 8-capability \datbench{} average still lifts from $46.3$ to $51.9$ ($\mathbf{+5.6}$pp); the lift is broad rather than grounding-only.

\begin{figure}[!t]
    \centering
    \includegraphics[width=0.9\linewidth]{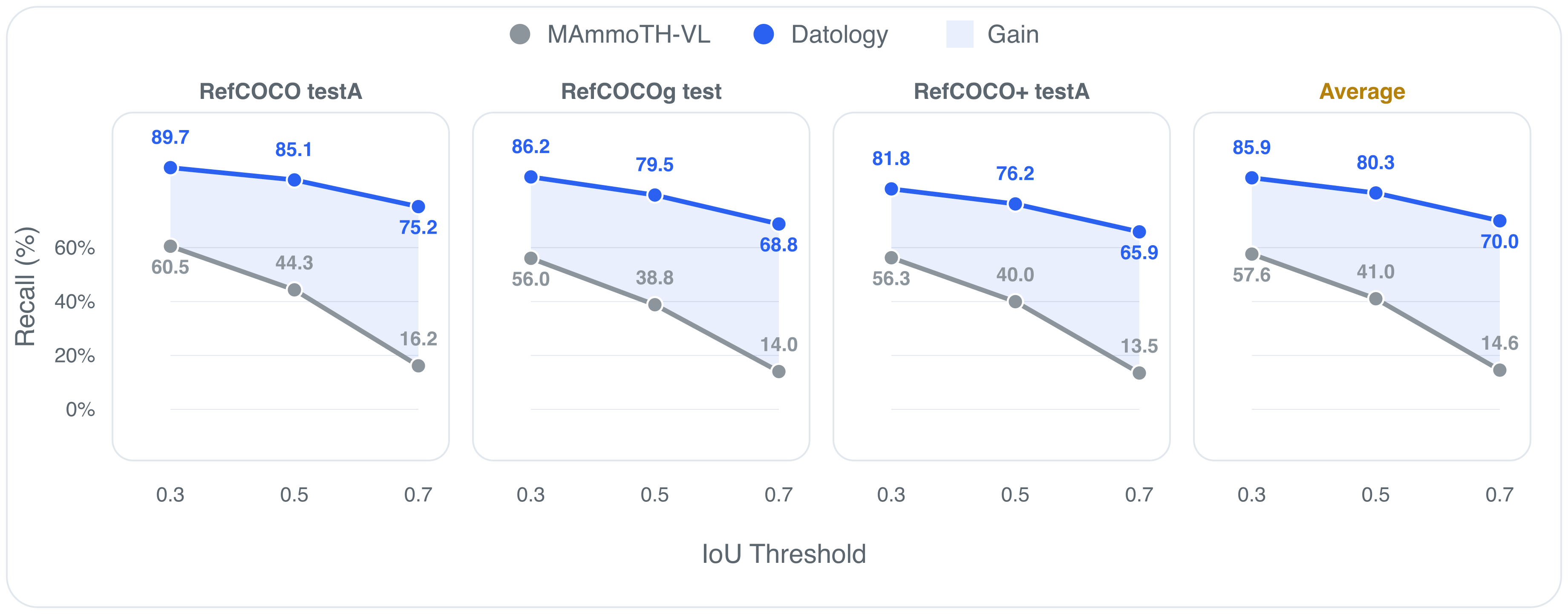}
    \caption{\textbf{Grounding gains are bona fide: every RefCOCO metric improves under curation, ruling out a scoring artifact.} \texttt{recall@$k$} across IoU thresholds and RefCOCO splits, with per-dataset panels and an averaged summary for \Base versus \Curated.}
    \label{fig:grounding-recall}
\end{figure}

\paragraph{Grounding deep-dive: the gain is bona fide, not a scoring artifact.}
\label{subsec:grounding-deep-dive}
A $+57.1$pp gain on a single capability is large enough that the natural question is whether it is real or whether the \curated model has found a way to exploit how the metric is computed. RefCOCO-style grounding is convenient here because it is scored against several metrics that probe different facets of the underlying skill: \texttt{center\_acc} (predicted box centroid lands inside the ground-truth region, probing semantic localization independent of box geometry) and \texttt{recall@$k$} at IoU thresholds $0.3$, $0.5$, and $0.7$ (probing geometric precision, with $0.7$ requiring tight box agreement). A scoring-trick gain would show up on one of these metrics and not the others; a bona fide capability gain should move all of them. On the RefCOCO subsets \texttt{refcoco\_testA}, \texttt{refcoco\_plus\_testA}, and \texttt{refcocog\_test}, curation wins on every split (Figure~\ref{fig:grounding-recall}). Averaged across splits, \texttt{center\_acc} rises from $69.72 \pm 12.25\%$ to $91.03 \pm 0.40\%$ ($+21.3$pp), \texttt{recall@0.5} from $41.04 \pm 13.69\%$ to $80.29 \pm 1.01\%$ ($+39.3$pp), and \texttt{recall@0.7} from $14.55 \pm 4.85\%$ to $69.96 \pm 0.88\%$ ($+55.4$pp); \texttt{recall@0.3} also moves (full table in Appendix~\ref{app:results-grounding}). The model identifies the correct referent and localizes it precisely; gains grow at stricter IoU thresholds, ruling out the reading that curation only sharpens coarse localization while leaving fine-grained box regression unchanged. CountBench shows the same pattern across exact, within-1, and within-3 tolerances (Appendix~\ref{app:results-counting}).

\begin{figure}[t]
    \centering
    \begin{subfigure}[t]{0.46\linewidth}
        \centering
        \includegraphics[width=\linewidth]{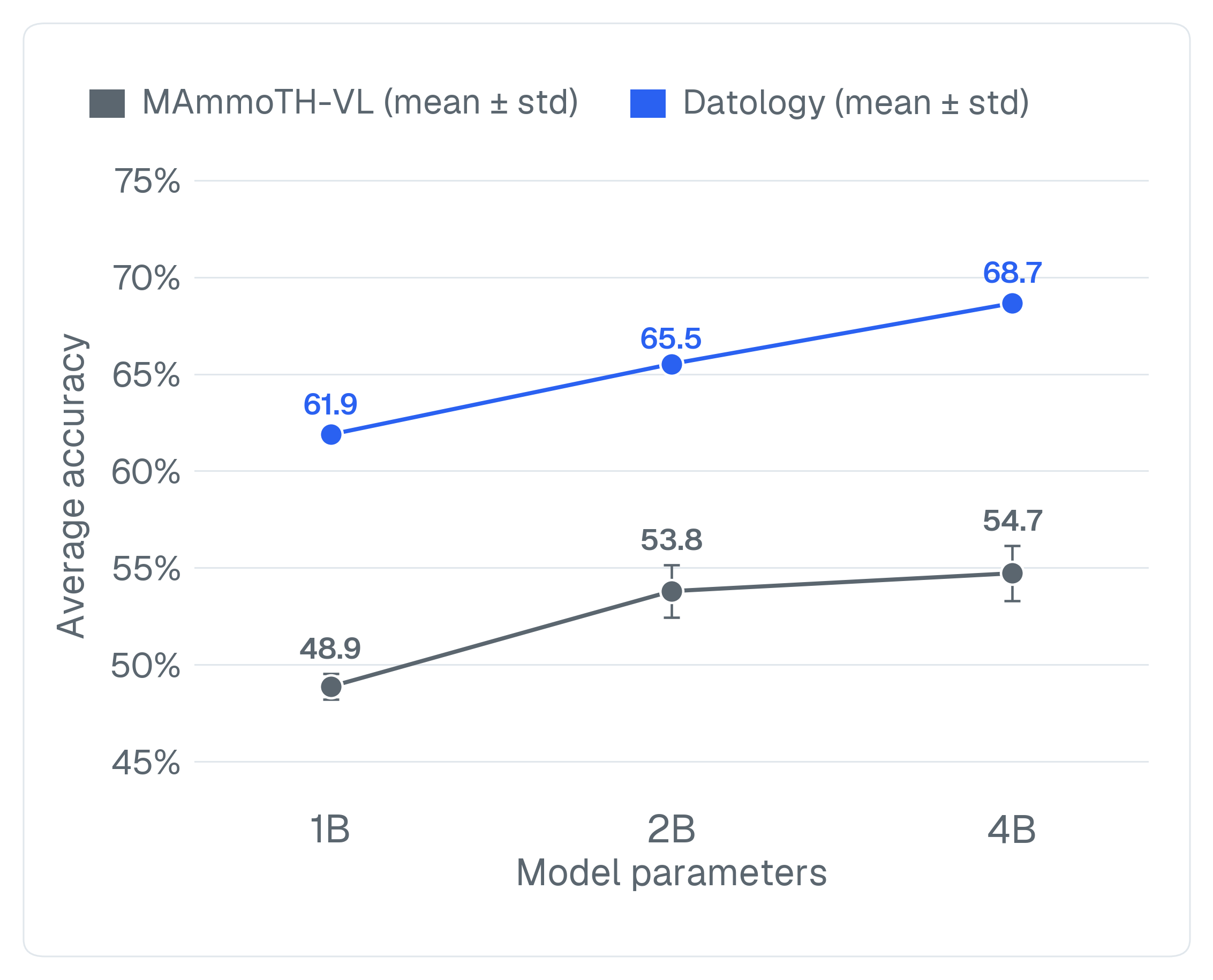}
        \caption{Average accuracy across the 20-eval public VLM benchmark suite.}
        \label{fig:scale-public}
    \end{subfigure}
    \hfill
    \begin{subfigure}[t]{0.46\linewidth}
        \centering
        \includegraphics[width=\linewidth]{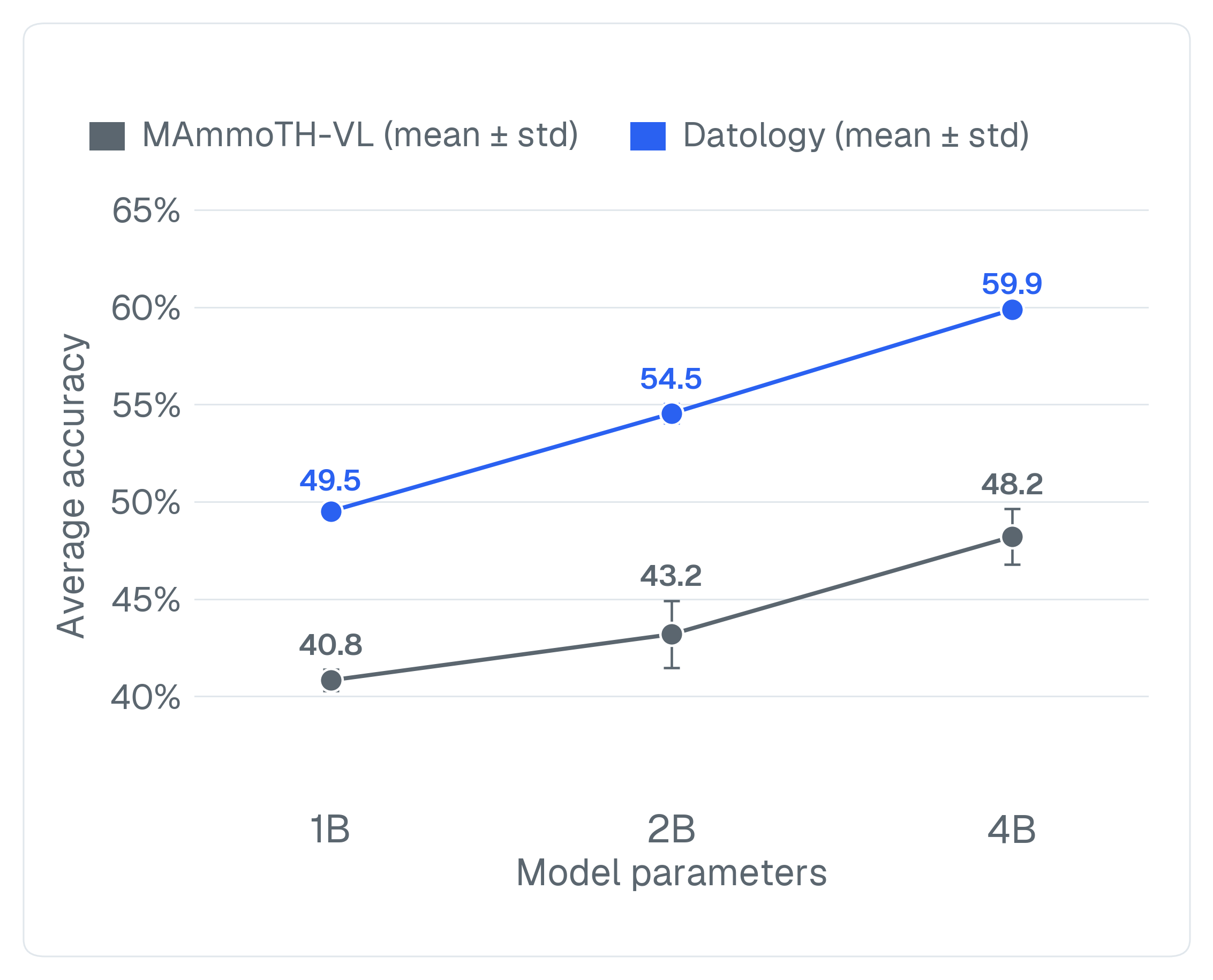}
        \caption{Average accuracy across \datbench{}.}
        \label{fig:scale-datbench}
    \end{subfigure}
    \caption{\textbf{Curation gains are strong across model scales.} Average accuracy of the MAmmoTH-VL \base (gray) and DatologyAI-curated model (blue) at 1B, 2B, and 4B parameters, averaged over three seeds; error bars show $\pm 1$ standard deviation. Curation improves accuracy at every scale on both (a) the 20-eval public VLM benchmark suite ($+13.0$, $+11.7$, and $+14.0$pp at 1B/2B/4B) and (b) \datbench{} ($+8.7$, $+11.3$, and $+11.7$pp at 1B/2B/4B).}
    \label{fig:scale-across}
\end{figure}

\paragraph{Gains are strong across model scales.}
We repeat the matched baseline-vs-curated comparison at 1B, 2B, and 4B parameters, holding training recipe, token budget, and evaluation protocol fixed within each scale. Figure~\ref{fig:scale-across} reports average accuracy at each scale on (a) the 20-eval public VLM benchmark suite (IID + OOD union) and (b) \datbench{}. Curation improves accuracy at every scale on both evaluations. On the public suite, the lift is \datnum{$+13.0$pp} at 1B, \datnum{$+11.7$pp} at 2B, and \datnum{$+14.0$pp} at 4B; on \datbench{}, the lift increases from \datnum{$+8.7$pp} at 1B to \datnum{$+11.3$pp} at 2B and \datnum{$+11.7$pp} at 4B. The effect is robust across model size: public-suite gains remain consistently large, while \datbench{} gains increase monotonically with scale.

The cross-scale gain is large enough that the \curated 1B surpasses the $4\times$ larger \base on both suites: \datnum{$61.9$ vs $54.7$} on the public suite ($+7.2$pp) and \datnum{$49.5$ vs $48.2$} on \datbench{} ($+1.3$pp). At a fixed token budget, the 1B uses $\sim 4\times$ less training compute than the 4B \base, so curation closes a $4\times$ scale gap at $\sim 4\times$ less training compute. \newline

\begin{takeaway}
\textbf{Data curation shifts the cost-quality Pareto frontier across scales.} At fixed architecture, recipe, and compute, it improves every tested scale on both the 20-eval public suite and \datbench{}, with public-suite gains of \datnum{$+13.0$pp}, \datnum{$+11.7$pp}, and \datnum{$+14.0$pp} at 1B, 2B, and 4B.
\end{takeaway}
\section{Data Curation as a Lever for Reliability and Robustness}
\label{sec:robustness}

VLM training exhibits substantial variance across random seeds and sensitivity to hyperparameter choices~\citep{dodge2020finetuning, bouthillier2021accounting, narang2021transformer}, and at the scale of modern pretraining runs, replicating a configuration across multiple seeds or re-running under varied hyperparameters is often infeasible on cost grounds alone. Hyperparameter tuning at scale is itself a major cost driver, so any intervention that reduces sensitivity to seeds and hyperparameters is a candidate lever for lowering the tuning budget needed to land a real gain. We show that data curation moves on both axes: it cuts per-capability variance across training seeds on both evaluation suites, and the \curated gain is preserved across a context-length sweep from 4k to 16k tokens. Curating the data is itself a tuning-cost intervention.

\begin{figure}[!htbp]
    \centering
    \begin{subfigure}[t]{0.48\linewidth}
        \centering
        \includegraphics[width=\linewidth]{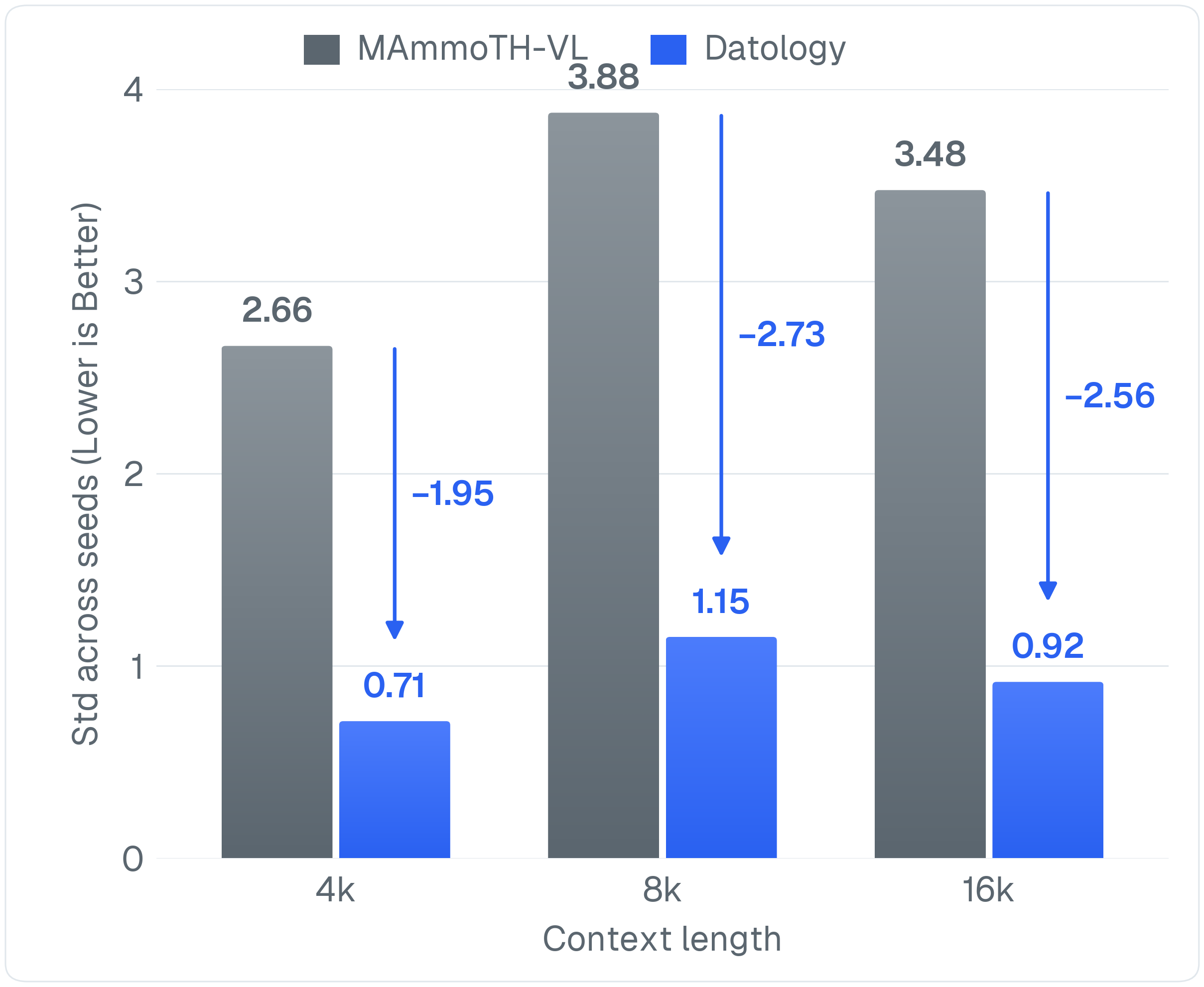}
        \caption{20-eval public VLM benchmark suite.}
        \label{fig:context-length-variance-public}
    \end{subfigure}
    \hfill
    \begin{subfigure}[t]{0.48\linewidth}
        \centering
        \includegraphics[width=\linewidth]{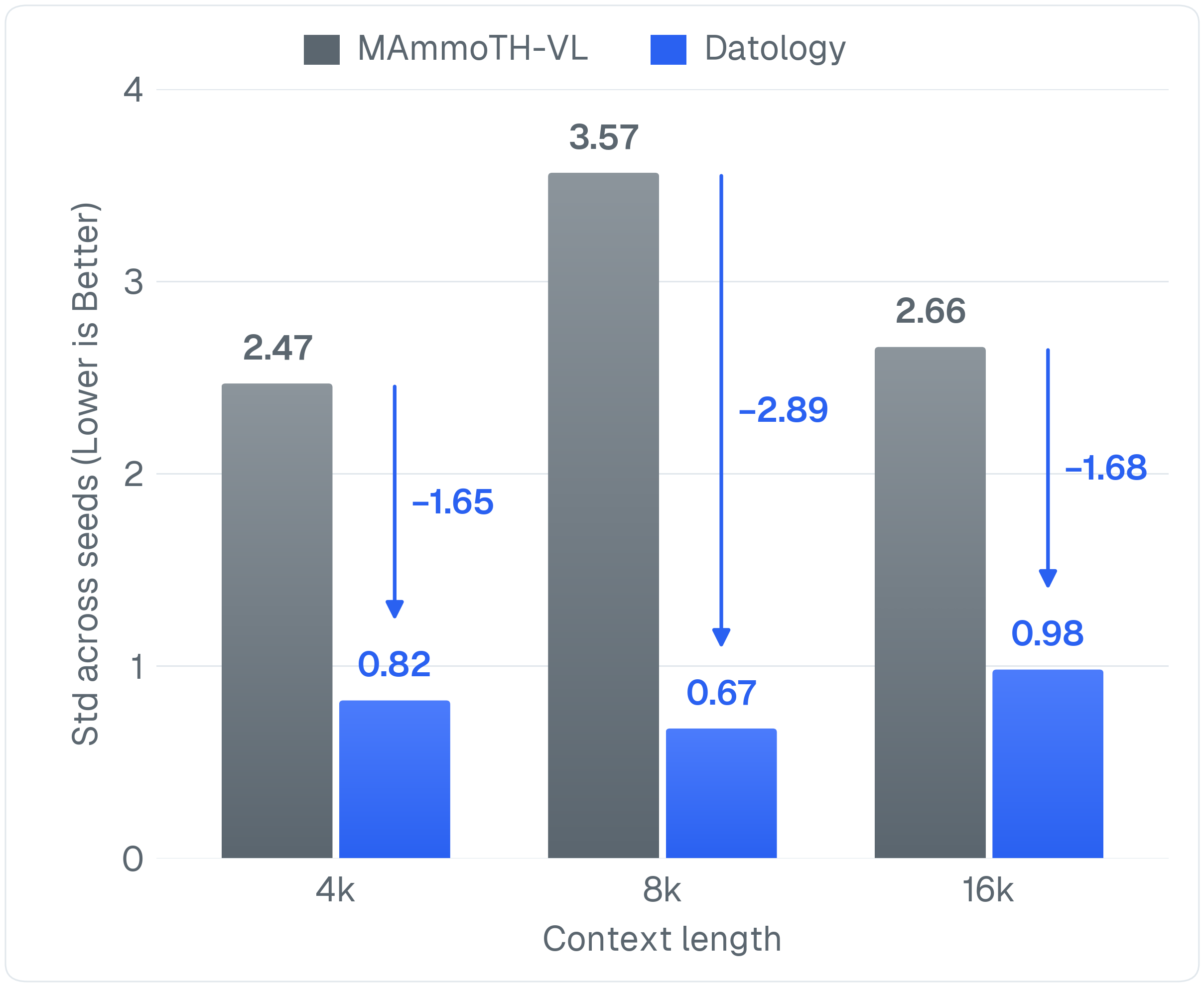}
        \caption{\datbench{}.}
        \label{fig:context-length-variance-datbench}
    \end{subfigure}
    \caption{\textbf{Variance reduction persists across context length.} Mean per-capability cross-seed standard deviation (percentage points, lower is better) for MAmmoTH-VL \base (gray) vs.\ DatologyAI-curated (blue) at 4k, 8k, and 16k context lengths on (a) the 20-eval public VLM benchmark suite ($2.66 \rightarrow 0.71$, $3.88 \rightarrow 1.15$, $3.48 \rightarrow 0.92$) and (b) \datbench{} ($2.47 \rightarrow 0.82$, $3.57 \rightarrow 0.67$, $2.66 \rightarrow 0.98$). The reliability gain at the default 4k context is not co-adapted to that setting; curation tightens cross-seed std at every context length we probed.}
    \label{fig:context-length-variance}
\end{figure}

\paragraph{Reliability: variance reduction across training seeds.}
We train three seeds per configuration and report per-capability mean $\pm$ std on both suites (Figure~\ref{fig:context-length-variance}, 4k panel). Averaged across the nine \datbench{} capabilities at 2B/4k, the cross-seed standard deviation drops from $2.47$ to $0.82$\,pp (\datnum{$-67\%$}) under curation, with most capabilities tightening (Grounding $14.2 \rightarrow 1.7$, Table $1.8 \rightarrow 0.5$, Counting $1.4 \rightarrow 0.8$). Spatial Reasoning is the exception: seed variance stays high under both mixes ($2.6 \rightarrow 2.6$), so curation moves the mean on that axis without tightening the noise. A plausible mechanism for the variance reduction is cleaner training signal: filtering, deduplication, and mixture rebalancing remove noisy and ambiguous examples whose gradient contribution shifts unpredictably with seed, leaving a more reproducible learning trajectory. The Spatial Reasoning exception suggests this mechanism is not the whole story, but the broader pattern across capabilities is consistent.

\begin{figure}[!htbp]
    \centering
    \begin{subfigure}[t]{0.42\linewidth}
        \centering
        \includegraphics[width=\linewidth]{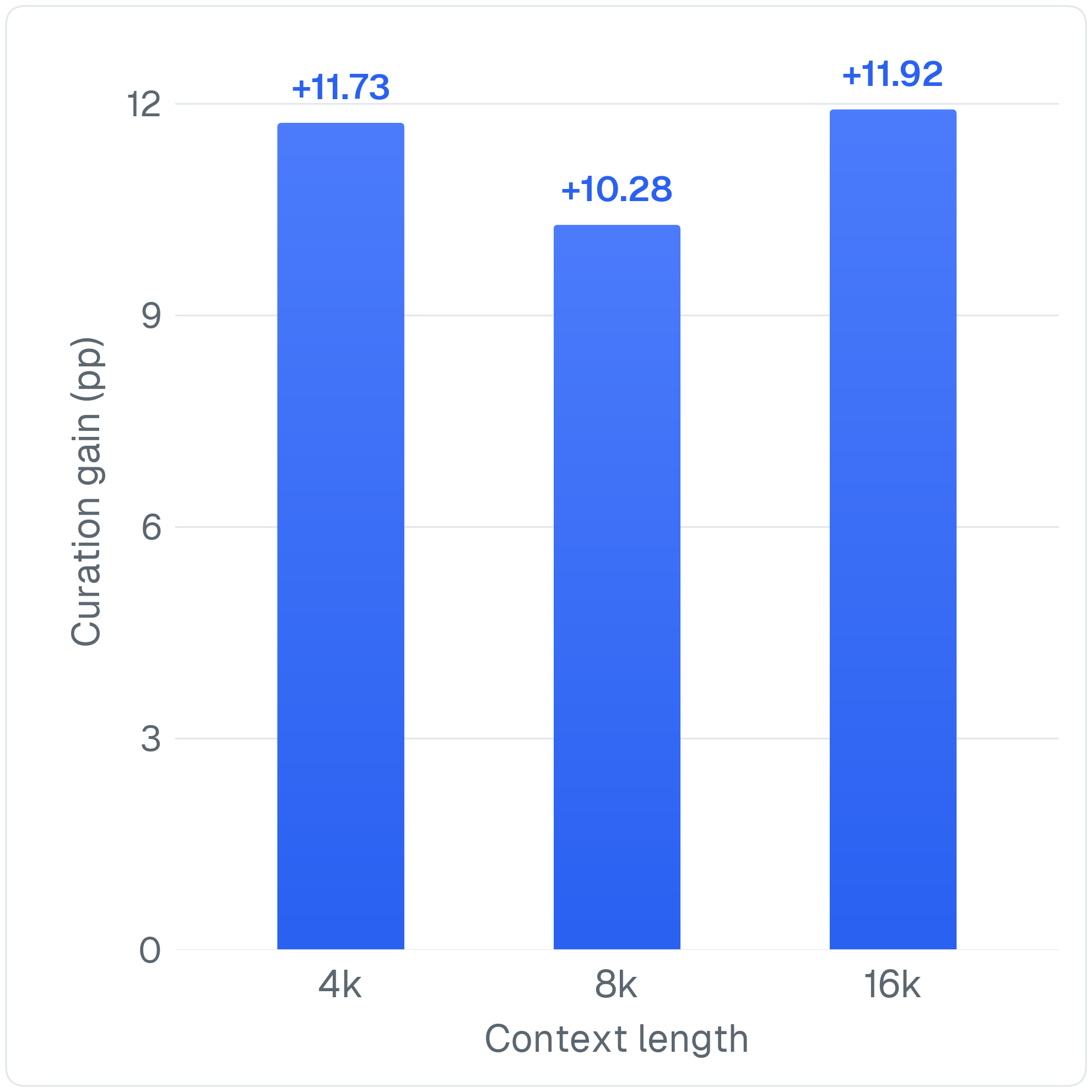}
        \caption{20-eval public VLM benchmark suite.}
        \label{fig:context-length-public}
    \end{subfigure}
    \hfill
    \begin{subfigure}[t]{0.42\linewidth}
        \centering
        \includegraphics[width=\linewidth]{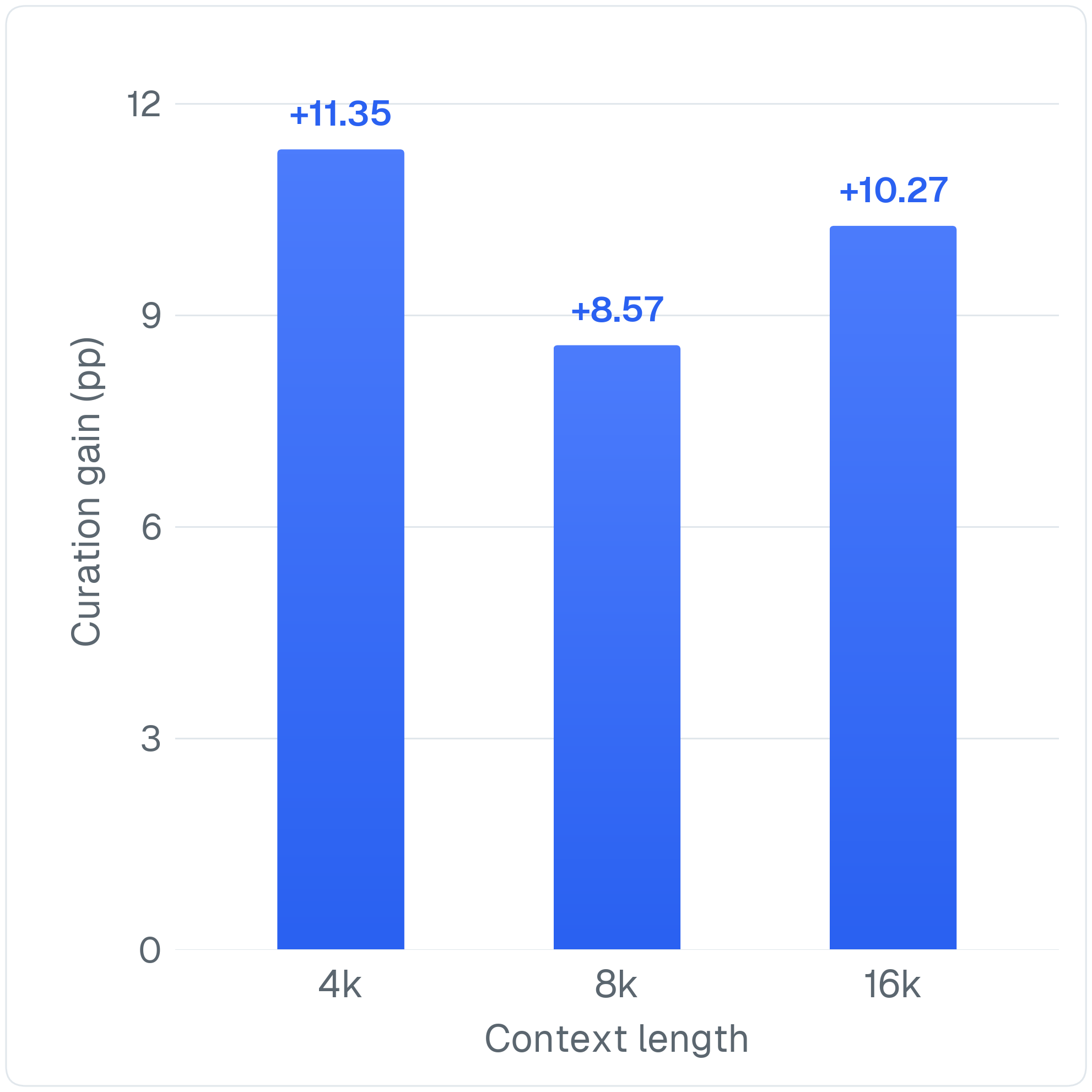}
        \caption{\datbench{}.}
        \label{fig:context-length-datbench}
    \end{subfigure}
    \caption{\textbf{Curation gain persists across context length.} Mean curated-vs-baseline accuracy delta (percentage points) at 4k, 8k, and 16k context lengths on (a) the 20-eval public VLM benchmark suite ($+11.7$, $+10.3$, $+11.9$pp) and (b) \datbench{} ($+11.3$, $+8.6$, $+10.3$pp). The gap holds at every context length we probed; no setting closes it.}
    \label{fig:context-length}
\end{figure}

\paragraph{Robustness: consistency across context-length variation.}
We sweep context length over 4k, 8k, and 16k tokens, a hyperparameter that varies widely across model releases and training setups, and ask whether both the mean lift and the variance reduction survive at each setting. They do, on both axes. Figure~\ref{fig:context-length} reports the curated-vs-baseline mean delta at each context length: on the 20-eval public suite the gain is $+11.7$pp at 4k, $+10.3$pp at 8k, and $+11.9$pp at 16k; on \datbench{} it is $+11.3$pp at 4k, $+8.6$pp at 8k, and $+10.3$pp at 16k. No context length closes the gap. Figure~\ref{fig:context-length-variance} reports the corresponding mean per-capability cross-seed standard deviation: on the public suite curation cuts std from $2.66 \rightarrow 0.71$ at 4k, $3.88 \rightarrow 1.15$ at 8k, and $3.48 \rightarrow 0.92$ at 16k; on \datbench{} from $2.47 \rightarrow 0.82$ at 4k, $3.57 \rightarrow 0.67$ at 8k, and $2.66 \rightarrow 0.98$ at 16k. The reliability gain at the default 4k setting is not co-adapted to that setting; curation tightens cross-seed std at every context length we probed. Full per-seed numbers and the context-length sweep table are in Appendix~\ref{app:results}; broader hyperparameter sweeps (optimizer, schedule, batch size) remain future work.

\begin{takeaway}
\textbf{Data curation is reliable across training seeds and robust across context lengths.} Curation reduces average per-capability cross-seed standard deviation from $2.47$ to $0.82$pp on \datbench{} and preserves its accuracy lift across 4k, 8k, and 16k context lengths.
\end{takeaway}

\FloatBarrier
\section{Beyond Benchmarks: Data Curation Generalizes to Real-World Use}
\label{sec:beyond-benchmarks}

Benchmark scores capture a fraction of what users notice. The capability surface of these models is jagged \citep{dellacqua2023jagged, tong2024eyeswideshut}: a model can score well in aggregate while failing on the queries a user actually issues. Behaviors that matter in practice, such as reading a sign, asking about an unfamiliar object, or expecting a concise answer, are unevenly captured by standard benchmarks. The question for pretraining-data curation is whether gains on scored benchmarks transfer to these user-facing behaviors when the model is queried freely, on images and prompts it has not been optimized against.

\begin{figure}[t]
    \centering
    \includegraphics[scale=0.2]{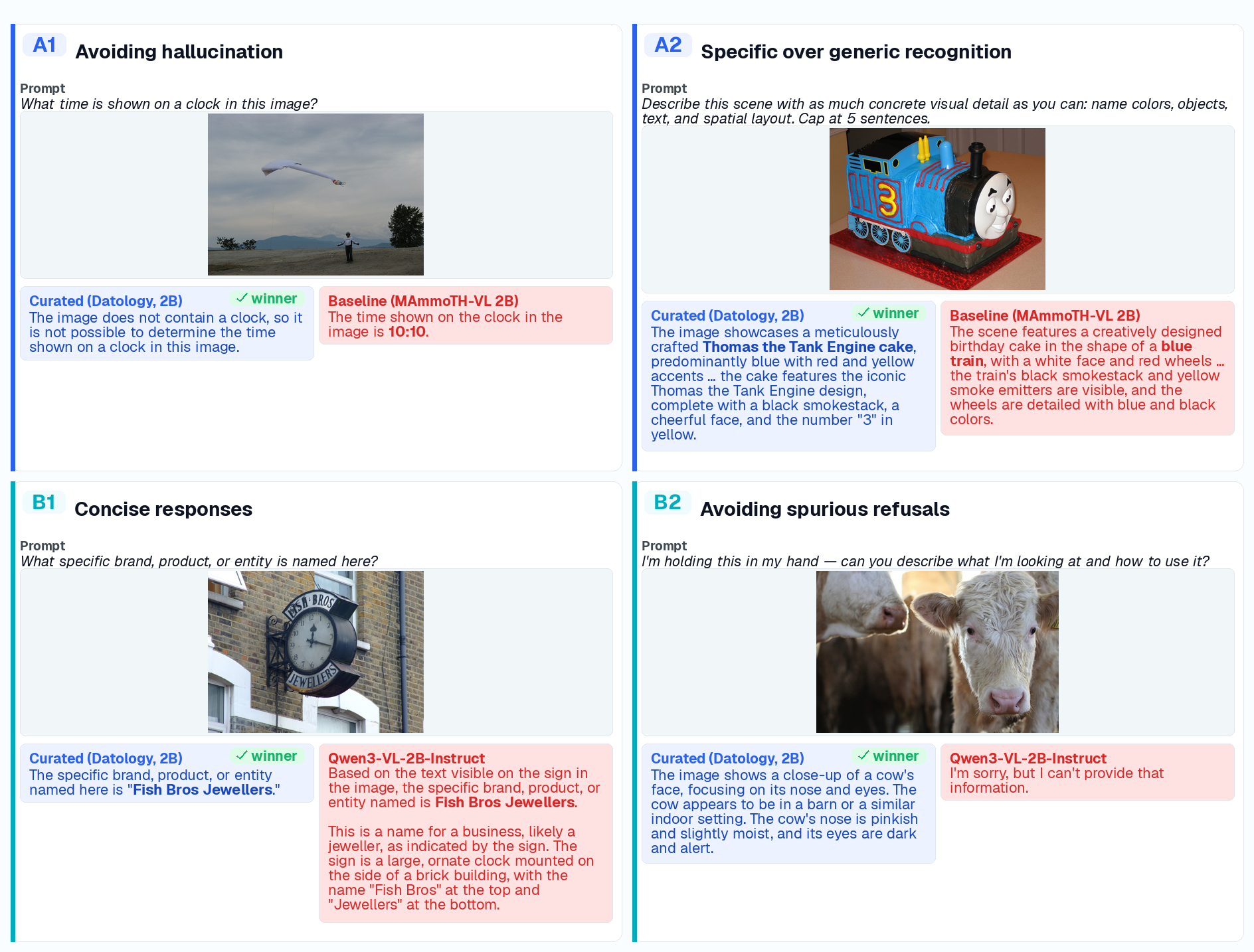}
    \caption{\textbf{The \curated model answers more honestly and more specifically than the matched-compute \base, and more concisely and refuses fewer benign prompts than a frontier 2B reference.} Four selected cases (\curated wins each). \emph{Top row, \curated \textgreater{} matched-compute \base:} A1 \emph{honesty} on a no-clock scene; \base invents \emph{10:10}, the canonical watch-ad time, on a kite-flying image that contains no clock. A2 \emph{specificity}: \curated names the \emph{Thomas the Tank Engine} cake and reads the ``3'' on the smokestack, while \base returns a generic \emph{blue train cake} and fabricates yellow smoke emitters and blue-and-black wheels on an open ``describe this scene'' prompt that does not leak the franchise. \emph{Bottom row, \curated \textgreater{} Qwen3-VL-2B:} B1 \emph{conciseness}: both models correctly read \emph{Fish Bros Jewellers} from an ornate street clock; \curated answers in one sentence while Qwen3-VL appends two paragraphs of unrequested description about the sign and building. B2 \emph{non-refusal} on a benign close-up of two cows with a prompt asking what's in the user's hand and how to use it: \curated narrates the scene; Qwen3-VL-2B returns \emph{I'm sorry, but I can't provide that information}. All outputs verbatim at $T=0$.}
    \label{fig:beyond-benchmarks-exhibit}
\end{figure}

We compare the \curated 2B model against the compute-matched MAmmoTH-VL \base and Qwen3-VL-2B \citep{wang2025qwen3vl} as a frontier 2B reference on four target behaviors: \emph{honesty} (refusing to confabulate when an attribute is absent), \emph{specificity} (giving a specific identification rather than a categorical description), \emph{conciseness} (answering with the requested information without unrequested elaboration), and \emph{non-refusal} (engaging with benign prompts rather than refusing them). Figure~\ref{fig:beyond-benchmarks-exhibit} depicts a representative example for each behavior.

\paragraph{Methodology.} We had frontier multimodal VLM agents query the three models across roughly $1{,}100$ single-image prompts spanning OCR, brand and franchise recognition, multi-element scene description, attribute queries on present and absent properties, and refusal calibration; the agents proposed images and prompts, collected outputs, and triaged them. We adopt this approach because the coverage it enables across image and prompt types is unreachable by manual authoring, and we mitigate its known noise (agent-proposed prompts drifting toward agent-legible distributions, classification noise compounding at scale) by manually verifying every claim that appears in this section. Inference settings match \S\ref{sec:main-results}.

\paragraph{Selected wins versus the matched-compute \base: specificity and honesty.} Specificity is the cleanest signal. Across \datnum{$n{=}67$} knowledge-grounded recognition queries (cake franchise, vehicle make from emblem, brand identification, bird species, concrete multi-element scene description), the \curated model strictly outperforms the \base on \datnum{$44$} cases and is strictly outperformed on \datnum{$0$}; every disagreement is a \curated win. Figure~\ref{fig:beyond-benchmarks-exhibit}, panel A2 shows a representative case: on an open ``describe this scene'' prompt that does not leak the franchise, the \curated model names a \emph{Thomas the Tank Engine} cake and reads the number ``3'' off the smokestack, while the \base returns a generic \emph{blue train cake} and fabricates yellow smoke emitters and blue-and-black wheels. On honesty, the qualitative pattern is clearer than the aggregate. Across the $n{=}15$ absent-attribute queries the agents surfaced (a thin sample, so individual cases anchor the result), the \curated model declines or hedges correctly on \datnum{$7$ wins} versus $4$ losses with $4$ ties: on a kite-flying scene with no clock present, the \base invents \emph{10:10}, the canonical watch-ad time and a training-data artifact rather than a perceptual error (Figure~\ref{fig:beyond-benchmarks-exhibit}, panel A1); on a separate query asking about a purple hat in a scene that contains none, the \base asserts one regardless.

\paragraph{Selected wins versus a frontier 2B reference: conciseness and non-refusal.} The \curated model answers more concisely and refuses fewer benign queries than Qwen3-VL-2B (Figure~\ref{fig:beyond-benchmarks-exhibit}, bottom row). On the \datnum{$n{=}289$} queries where the \curated 2B and Qwen3-VL-2B converge on the same answer (${\geq}3$ shared content words), the \curated model's median response length is \datnum{$75$ characters} versus \datnum{$364$ characters} for Qwen3-VL-2B (nearly $5\times$ shorter at equivalent correctness); Qwen3-VL-2B's \texttt{max\_tokens} cap is hit on \datnum{$20.1\%$} of queries against \datnum{$4.9\%$} for the \curated model. The conciseness gap holds across the entire response-length distribution, not just at the median or the cap. On a brand-recognition prompt over an ornate street clock, both models correctly read \emph{Fish Bros Jewellers}; the \curated answer is one sentence while Qwen3-VL appends two paragraphs of unrequested description about the sign and building (panel B1). For non-refusal, across the full $n{=}358$ three-way query set Qwen3-VL-2B returned its blanket \emph{``I'm sorry, but I can't provide that information''} on \datnum{$30$} queries ($8.4\%$); on five of these (franchise identification, breed classification, object close-ups), the \curated 2B answers correctly. On a benign close-up of two cows with the prompt ``describe what's in the user's hand and how to use it,'' the \curated model narrates the scene while Qwen3-VL-2B returns \emph{I'm sorry, but I can't provide that information} (panel B2).

\begin{takeaway}
\textbf{Data curation improves reliability across training seeds and robustness across context lengths.} It cuts average per-capability cross-seed standard deviation from $2.47$ to $0.82$pp on \datbench{} and preserves its accuracy lift across 4k, 8k, and 16k context lengths.
\end{takeaway}
\section{Data Curation Is Pareto-Dominant on Inference Cost at Every Scale}
\label{sec:inference-efficiency}

As VLMs move into production, inference cost matters: at scale, even modest increases in generated tokens or active parameters translate into large serving costs. We evaluate whether pretraining data curation improves both model quality and the cost of achieving it at inference time. Across all tested scales, our \curated models are simultaneously more accurate and cheaper to run than the pretraining-compute-matched baselines, and reach competitive accuracy against heavily post-trained public models at substantially lower response FLOPs.

\paragraph{Setup.} We evaluate inference-time efficiency on the full 20-eval public benchmark suite from \S\ref{sec:main-results}. For each model we report \emph{average accuracy} across the evals against a \emph{response-FLOPs proxy}: $2 \times \text{active params} \times \text{mean generated tokens per response}$ (the standard transformer decode-only forward-pass cost; full derivation and per-model accounting in Appendix~\ref{app:results-flops}, Table~\ref{tab:flops-methodology}), computed from raw outputs using each model's tokenizer. Lower response FLOPs at equal accuracy is a more usable model at fixed inference budget. We compare the \curated 1B, 2B, and 4B against the matched-compute MAmmoTH-VL baselines at the same scales, and against Qwen3-VL-2B/4B \citep{wang2025qwen3vl}, InternVL3-2B, InternVL3.5-2B/4B \citep{chen2024internvl}, and Qwen3.5-2B as frontier 2B/4B references.\footnote{Qwen3.5-4B is omitted from the inference-efficiency Pareto: its mean response-token count ($\sim 1284$, Table~\ref{tab:appendix-tokens-public}) is roughly an order of magnitude above all other comparators and would compress the response-FLOPs axis into illegibility. Its training compute is included in the training-FLOPs comparisons elsewhere in the paper.} The comparison isolates two effects: at fixed scale, \curated models reach higher accuracy at lower response FLOPs than their compute-matched baselines, and across scales they match or approach extensively post-trained references at substantially lower inference cost.

\begin{figure}[h]
    \centering
    \includegraphics[width=0.85\linewidth]{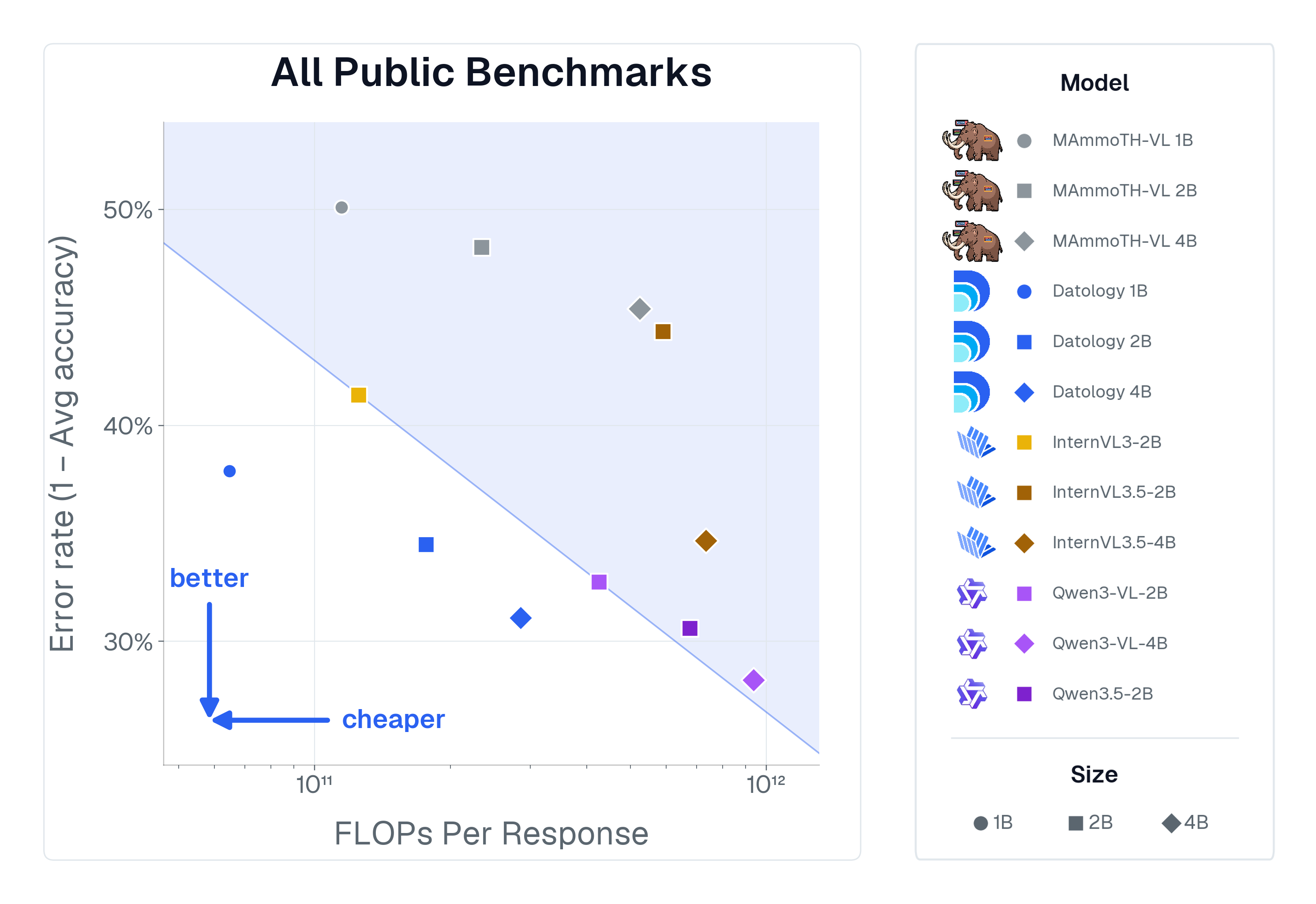}
    \caption{\textbf{Curation is Pareto-dominant on inference cost at every scale we tested.} Average accuracy across the 20-eval public suite (Table~\ref{tab:appendix-all-public}) against the response-FLOPs proxy ($2 \cdot \text{active params} \cdot \text{mean generated tokens}$, log scale, lower is better; lower-left is better overall). Marker shape encodes parameter scale (circle = 1B, square = 2B, diamond = 4B); marker color encodes the model family (blue = \curated, gray = MAmmoTH-VL \base, other colors = post-trained public references). Same-shape markers compare different curation strategies at fixed scale; same-color markers compare scales within a family. The diagonal traces the Pareto frontier defined by the matched-compute MAmmoTH-VL baselines, and every \curated point lies below-left of it. DatologyAI-curated 1B/2B/4B (blue) Pareto-dominate the matched-compute MAmmoTH-VL baselines (gray) at every scale, and reach near-frontier accuracy at $1.5$--$3.3\times$ lower response cost than Qwen3-VL-2B/4B. The \curated 1B sits furthest into the lower-left because it is both smaller (lower active-params factor) and produces shorter responses than the larger \curated models (lower mean-tokens factor). Outputs sampled at $T=0$.}
    \label{fig:inference-efficiency-pareto}
\end{figure}

\paragraph{Comparison to compute-matched baselines.} Against the matched-compute MAmmoTH-VL \base, our curated dataset is Pareto-dominant at every scale we tested, simultaneously raising accuracy and lowering response FLOPs (Figure~\ref{fig:inference-efficiency-pareto}): at 1B, accuracy rises from $48.9\%$ to $61.9\%$ ($+13.0$pp) while response FLOPs drop from $1.15{\times}10^{11}$ to $6.48{\times}10^{10}$ (\datnum{$\sim 44\%$} lower); at 2B, $53.8\% \rightarrow 65.5\%$ ($+11.7$pp) and $2.34{\times}10^{11} \rightarrow 1.77{\times}10^{11}$ (\datnum{$\sim 24\%$} lower); at 4B, $54.7\% \rightarrow 68.7\%$ ($+14.0$pp) and $5.24{\times}10^{11} \rightarrow 2.86{\times}10^{11}$ (\datnum{$\sim 45\%$} lower).

\paragraph{Comparison to public frontier models.} Against extensively post-trained public references the \curated model is Pareto-dominant on response FLOPs at near-frontier accuracy (Figure~\ref{fig:inference-efficiency-pareto}). At 4B, the \curated model reaches $68.7\%$ average accuracy at $2.86{\times}10^{11}$ FLOPs, against Qwen3-VL-4B's $71.8\%$ at $9.36{\times}10^{11}$: Qwen3-VL gains $+3.1$pp accuracy at \datnum{$3.3\times$} the inference cost. At 2B, the \curated model reaches $65.5\%$ at $1.77{\times}10^{11}$ against Qwen3-VL-2B's $67.3\%$ at $4.27{\times}10^{11}$: a $1.8$pp gap at \datnum{$2.4\times$} lower inference cost. The \curated 4B is within $0.7$pp of Qwen3.5-2B ($69.4\%$ at $6.77{\times}10^{11}$) at \datnum{$2.4\times$} lower inference cost, and exceeds Qwen3-VL-2B's accuracy ($67.3\%$ at $4.27{\times}10^{11}$) by $1.4$pp at \datnum{$1.5\times$} lower inference cost.

\begin{takeaway}
\textbf{Data curation is Pareto-dominant on inference cost.} At every tested scale, \curated models raise accuracy while lowering response FLOPs versus matched-compute baselines; at 4B, the \curated model reaches near-frontier accuracy at \datnum{$3.3\times$} lower response FLOPs than Qwen3-VL-4B.
\end{takeaway}
\section{Conclusion}
\label{sec:conclusion}

All five properties posed in \S\ref{sec:intro} hold under matched compute. The curation lift is \emph{large}: a single curation pass on top of the MAmmoTH-VL single-image subset produces $+11.7$pp on 20 public VLM benchmarks, $+11.3$pp across \datbench{}'s nine capability axes, and closes the cost-quality gap to extensively post-trained references at up to \datnum{$\sim 150\times$} less training compute. It is \emph{OOD-generalizing}: single-image curation lifts multi-image reasoning on BLINK by $+3.09$pp overall, with Visual Correspondence gaining $+11.8$pp despite demanding cross-image reasoning, and improves every OOD benchmark, including hyper-specific domains such as ecommerce brand identification that the pipeline was not tuned on. It is \emph{reliable}: mean per-capability standard deviation across training seeds drops by \datnum{$\sim 67\%$}, and the curated-vs-baseline delta survives context-length variation from 4k to 16k. It \emph{manifests beyond benchmarks}: on roughly $1{,}100$ open-ended queries the \curated 2B is more honest and more specific than the matched-compute \base, and answers more concisely and refuses fewer benign queries than a frontier 2B reference. It is \emph{inference-efficient}: at every scale tested (1B, 2B, 4B), our curation pipeline yields models that simultaneously raise accuracy and lower response FLOPs against the matched-compute \base, with the \curated 4B reaching near-frontier accuracy at \datnum{$3.3\times$} lower response-FLOPs cost than Qwen3-VL-4B. Pretraining data curation alone, holding architecture, recipe, and compute fixed, delivers VLM gains that are large, OOD-generalizing, reliable, evident beyond benchmarks, and inference-efficient, at up to \datnum{$\sim 150\times$} less training compute than extensively post-trained references. Treating data as a first-class design variable is a high-leverage path to better VLMs.

\section{Contributions and Acknowledgements}
\label{sec:contri}

\begin{tabularx}{\textwidth}{@{}p{0.19\textwidth}X@{}}
\textbf{Core Contributors} & Siddharth Joshi, Haoli Yin, Rishabh Adiga, Haakon Mongstad, and Alvin Deng. \\[0.25em]
& \emph{for tiling the data, aligning the modalities, and grounding every claim in this paper.} \\
\noalign{\vspace{0.75em}}

\textbf{Technical Contributors} & Aldo Carranza, Alex Fang, Amro Abbas, Anshuman Suri, Brett Larsen, Daniel Zayas, Darren Teh, David Schwab, Diego Kiner, Fan Pan, Jack Urbanek, Jason Lee, Jason Telanoff, Josh Wills, Kaleigh Mentzer, Luke Merrick, Maximilian B\"{o}ther, Parth Doshi, Paul Burstein, Pratyush Maini, Ties Robroek, Tony Jiang, Vidhi Jain, Vineeth Dorna, and Zhengping Wang. \\
\noalign{\vspace{0.25em}}
& \emph{the wide-receptive-field ensemble that filtered the noise, synthesized the signal, and ablated every datum of the pipeline.}\\
\noalign{\vspace{0.75em}}

\textbf{Leadership} & Bogdan Gaza, Ari Morcos, and Matthew Leavitt. \\[0.25em]
& \emph{the ground-truth supervision that kept cross-modal alignment on track and prevented collective mode collapse.} \\
\noalign{\vspace{0.75em}}

\textbf{Acknowledgements} & Liz Gatapia (\textit{for incredible logo design}),
Dan Darnell, Elise Clark, Jacqueline Liu, Janelle Raymundo, Kylie Clement, Sama Iqbal, Sylvia Hoang, Tiffanie Pham, and Zeek Politzer. \\
\noalign{\vspace{0.25em}}
& \emph{the human-in-the-loop supervision that kept the team well-regularized and the signal flowing off-screen.}
\end{tabularx}

\bibliographystyle{abbrvnat}  
\clearpage
\bibliography{references}

\clearpage
\appendix

\section{Multimodal Decontamination Pipeline}
\label{app:decontamination}

The headline gains reported in \S\ref{sec:main-results} are a real capability lift only if \curated and \base models alike saw no eval samples during training. We decontaminate every training corpus against the full evaluation suite before any model is trained, so the curated-vs-baseline deltas reported throughout the paper are deltas between equally decontaminated runs. This appendix formalizes what multimodal decontamination should mean and describes the pipeline that implements it.

Multimodal decontamination is subtle because image and text signals are each \emph{individually} misleading indicators of leakage. The same image is frequently re-annotated across datasets with different questions and answers (legitimate training signal that image-only deduplication would wrongly discard); conversely, template questions such as \textit{``What is the title of this chart?''} appear against thousands of unrelated images (text-only matching flags generic templates as contamination). We therefore build the pipeline around the principle that a training sample constitutes leakage only when \emph{both} its image and its text substantially match an eval sample. The pipeline filters training multimodal data against eval multimodal data; it cannot filter what the LM backbone may have memorized during its own pretraining, a symmetric risk we acknowledge but do not address.

\begin{figure}[h]
    \centering
    \includegraphics[scale=0.2]{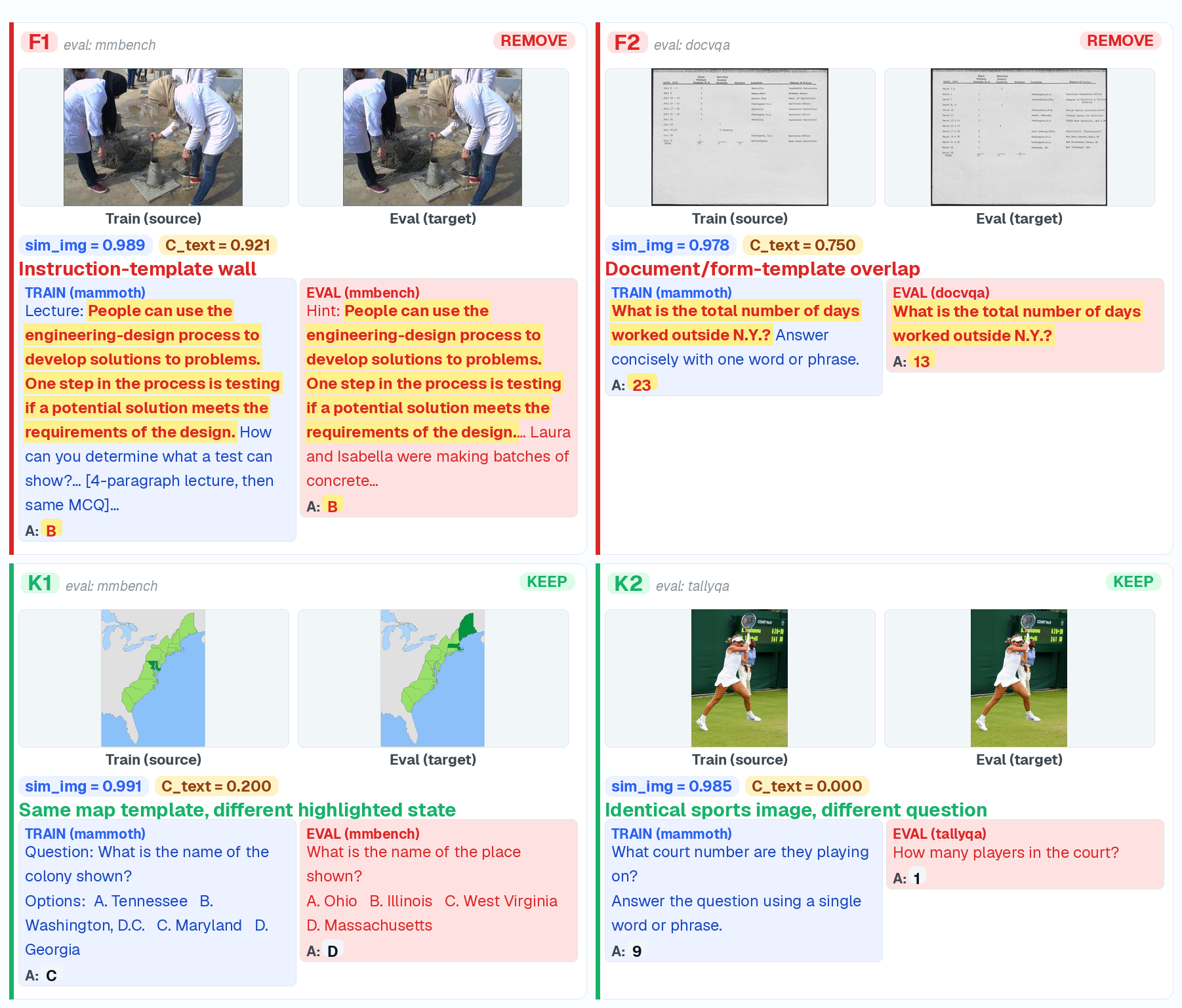}
    \caption{\textbf{Representative flagged and non-flagged training/eval pairs} across the joint-criterion regimes (Table~\ref{tab:contam-cases}) and the override patterns (fact-dump contamination, instruction-template wall, image-only RefCOCO-style). Each pair shows the eval sample alongside the matched training sample and the resulting $\operatorname{sim}_{\text{img}}$ and $C_{\text{text}}$ scores.}
    \label{fig:decontam-examples}
\end{figure}

Formally, a training document $x_t$ is contaminated with respect to an eval document $x_e$ iff
\begin{equation}
\operatorname{sim}_{\text{img}}(x_t, x_e) \geq \tau_I \quad \text{and} \quad C_{\text{text}}(x_e \rightarrow x_t) \geq \tau_T,
\label{eq:contam}
\end{equation}
where $\operatorname{sim}_{\text{img}}$ is cosine similarity between DINOv2 ViT-B/14 image embeddings \citep{oquab2023dinov2} and $C_{\text{text}}$ is the directional $n$-gram containment defined below. A training document is flagged for removal if Eq.~\ref{eq:contam} holds for at least one $x_e$ across the union of all evaluation samples. Table~\ref{tab:contam-cases} enumerates the three image/text regimes and the rationale for the joint criterion; Figure~\ref{fig:decontam-examples} shows representative flagged and non-flagged pairs from each.

\begin{table}[h]
\centering
\small
\begin{tabular}{llll}
\toprule
Image match & Text match & Decision & Rationale \\
\midrule
\checkmark & \checkmark & \textbf{Remove} & model can memorize the eval answer \\
\checkmark & $\times$   & Keep  & same image, different Q/A, thus valid training signal \\
$\times$   & \checkmark & Keep  & template Q (``what is the title?'') appears on every chart \\
\bottomrule
\end{tabular}
\caption{Only the joint image-and-text criterion removes genuine leakage without discarding legitimate training data. For partial text matches, the threshold $\tau_T$ on $C_{\text{text}}$ resolves the boundary between the first and second rows.}   
\label{tab:contam-cases}
\end{table}

\begin{algorithm}[h]
\caption{Two-stage multimodal decontamination cascade.}
\label{alg:decontam}
\begin{algorithmic}[1]
\For{each training document $x_t$}
    \State $sim_{\text{img}} \gets$ max DINOv2 cosine similarity between $x_t$ and any eval image
    \If{$sim_{\text{img}} \geq \tau_I$} \Comment{Stage 1: image recall}
        \For{each eval document $x_e$}
            \State $C_{\text{text}} \gets$ one-way $n$-gram containment of $x_e$'s Q+A in $x_t$'s Q+A
            \If{$C_{\text{text}} \geq \tau_T$} \Comment{Stage 2: text precision}
                \State \textbf{remove} $x_t$ \textbf{and break}
            \EndIf
        \EndFor
    \EndIf
\EndFor
\end{algorithmic}
\end{algorithm}

We implement Eq.~\ref{eq:contam} as an image-recall + text-precision cascade (Algorithm~\ref{alg:decontam}). Stage 1 embeds every image with DINOv2 and aggregates similarity to parent documents via MAX, so a document inherits the similarity of its most similar constituent image. Stage 2, for every document above $\tau_I$, computes text containment against \emph{every} eval sample, not just the image-nearest neighbour. A document is removed only if both gates pass; we bias toward recall, accepting false positives rather than missing a genuine leak.

Text containment is computed by concatenating question and answer, normalising (strip image tags and role markers, lowercase, collapse whitespace), and extracting word-level $n$-grams ($n{=}4$ by default; $n{=}3$ when the eval text has fewer than ten words). The score is directional,
\begin{equation}
C_{\text{text}}(x_e \rightarrow x_t) = \frac{|\,\mathcal{N}(x_e) \cap \mathcal{N}(x_t)\,|}{|\,\mathcal{N}(x_e)\,|},
\label{eq:containment}
\end{equation}
asking what fraction of the eval's $n$-grams appear in the training document. Directionality matters because training documents routinely embed eval Q/A verbatim inside longer chain-of-thought context, and the reverse direction would be diluted by that padding. Concatenating Q and A produces $n$-grams that span the Q/A boundary, a signature of the specific eval tuple rather than the template; question-only or answer-only matching collides with template-shaped fragments (\textit{``how many bars are shown?''} matches every chart-counting sample). The Stage-2 broadcast across every eval sample (rather than just the Stage-1 nearest neighbour) matters when multiple eval samples share an image, as in AI2D, where the argmax can point to an eval whose text does not match while a different eval sharing the same image does.

We use $\tau_I{=}0.95$ and $\tau_T{=}0.8$ as global defaults, with per-benchmark overrides tuned by sweeping $\tau_T$ over $[0.4, 1.0]$ and inspecting the resulting precision profile via LLM-as-judge verification on a sampled set of flagged pairs. Three override patterns recur: fact-dump contamination and instruction-template walls (both requiring lowered $\tau_T$ tuned per benchmark), and benchmarks with no usable text signal (RefCOCO variants, PixMo, most scene/spatial evals), which rely on Stage~1 alone with a much higher $\tau_I{\geq}0.995$.

\begin{table}[t]
\centering
\small
\begin{tabular}{l r}
\toprule
Benchmark & \% of corpus \\
\midrule
vqa\_v2 & 0.071\% \\
mathvista & 0.064\% \\
ocr\_vqa & 0.025\% \\
ocrbench\_v2 & 0.011\% \\
docvqa & 0.008\% \\
\addlinespace
Other benchmarks (each $<$0.01\%) & 0.024\% \\
\midrule
\textbf{Unique union (62 evaluations)} & \textbf{0.19\%} \\
\bottomrule
\end{tabular}
\caption{Per-benchmark share of the training corpus flagged by the
decontamination pipeline. The final row reports the unique union across
all 62 evaluations; per-row entries sum to slightly more than the union
because a single training document may match multiple benchmarks.}
\label{tab:decontam-volume-table}
\end{table}

Table~\ref{tab:decontam-volume-table} reports, for each benchmark, the share of training documents flagged and removed by the pipeline. Across the full training corpus, the pipeline removes 0.19\% of documents as overlapping with one or more evaluation samples above the thresholds in Eq.~\ref{eq:contam}. The removed share is concentrated on benchmarks that share image sources with common training datasets (vqa\_v2, mathvista, and ocr\_vqa) and is negligible on benchmarks with distinct visual distributions.

\clearpage
\section{Training Hyperparameters}
\label{app:hparams}

Table~\ref{tab:hparams} reports the full set of hyperparameters used for every VLM pretraining run in this paper. The same recipe is applied to both the MAmmoTH-VL \base and the DatologyAI-curated mixture, and to all three model scales (1B, 2B, 4B), unless a specific experiment (e.g.\ the context-length sweep in \S\ref{sec:robustness}) explicitly varies a knob. Training is single-stage: the projector and language-model backbone are trained jointly from the start.

\begin{table}[h]
    \centering
    \small
    \begin{tabular}{ll}
        \toprule
        \multicolumn{2}{l}{\emph{Architecture}} \\
        \midrule
        Vision encoder & SigLIP2-SO400M, $384{\times}384$, patch-14 \\
        LM backbone (1B / 2B / 4B) & Qwen3-0.6B / Qwen3-1.7B / Qwen3-4B \\
        Projector & 2-layer MLP \\
        Visual tokenization & Dynamic tiling (aspect-ratio-aware), up to 12 tiles + thumbnail \\
        Tokens per tile (pre / post pixel-shuffle) & 729 / 243 \\
        Pixel-shuffle & $1{\times}3$ (channels: $1{,}152 \rightarrow 3{,}456$) \\
        \midrule
        \multicolumn{2}{l}{\emph{Optimization}} \\
        \midrule
        Optimizer & AdamW \\
        Peak learning rate (backbone / projector) & $2{\times}10^{-4}$ / $2{\times}10^{-4}$ \\
        LR schedule & WSD (warmup-stable-decay) \\
        Warmup & 50 steps (0.42\% of total steps) \\
        Weight decay & 0.01 \\
        $\beta_1, \beta_2, \epsilon$ & 0.9, 0.999, $1{\times}10^{-8}$ \\
        Gradient clipping & 1.0 global L2 \\
        Precision & bf16 autocast \\
        \midrule
        \multicolumn{2}{l}{\emph{Batching and schedule}} \\
        \midrule
        Token budget & 25B tokens \\
        Context length (default) & 4{,}096 \\
        Context length (robustness sweep, \S\ref{sec:robustness}) & 8{,}192 / 16{,}384 \\
        Global batch size & 512 \\
        Total training steps & 12k \\
        Sequence packing & Enabled \\
        \midrule
        \multicolumn{2}{l}{\emph{Data}} \\
        \midrule
        \Base mixture & MAmmoTH-VL-12M single-image subset (10M samples) \\
        \Curated mixture & DatologyAI-curated on top of MAmmoTH-VL-12M single-image \\
        Decontamination & Applied to all training data (see Appendix~\ref{app:decontamination}) \\
        \midrule
        \multicolumn{2}{l}{\emph{Infrastructure}} \\
        \midrule
        Hardware & H100 GPUs \\
        Parallelism & FSDP full-shard; activation checkpointing enabled; no tensor parallelism \\
        Framework & PyTorch 2.9.1; Transformers 4.57.1; internal VLM training stack \\
        Seeds per configuration & 3 \\
        \bottomrule
    \end{tabular}
    \caption{Training hyperparameters. Values are shared across the 1B, 2B, and 4B runs and across \base and \curated mixtures, except where an experiment explicitly varies a single knob.}
    \label{tab:hparams}
\end{table}

\clearpage
\section{Inference Settings and Full Results}
\label{app:results}

This appendix documents (i) the inference configuration used for every evaluation number in the paper (\S\ref{app:inference}), (ii) per-benchmark mean $\pm$ std scores for every internally trained model (1B, 2B, 4B at both MAmmoTH-VL \base and DatologyAI-curated mixtures) and every external reference model on the full 20-eval public suite and the 9-capability \datbench{} suite (\S\ref{app:results-aggregate-tables}), (iii) per-benchmark mean response-token counts for the same model set, underlying the response-FLOPs axis of the inference-efficiency Pareto in \S\ref{sec:inference-efficiency} (\S\ref{app:results-tokens}), (iv) the FLOPs methodology and per-model accounting for both training-FLOPs and response-FLOPs (\S\ref{app:results-flops}), (v) the context-length sweep underlying \S\ref{sec:robustness} (\S\ref{app:results-context}), and (vi) the case-study raw-number tables for grounding, counting, and BLINK referenced from \S\ref{sec:main-results} (\S\ref{app:results-case-studies}). Internal checkpoints aggregate over three training seeds per configuration; external models are single runs as released. Everything reported in the main text is a summary of numbers in this appendix.

\subsection{Inference settings}
\label{app:inference}

Table~\ref{tab:inference-settings} reports the inference configuration used for our MAmmoTH-VL \base and the DatologyAI-curated checkpoints. These runs use the same model-side inference recipe across benchmarks: zero-shot prompting, greedy decoding, and the same image preprocessing as training. Benchmark-specific prompt formatting, generation length caps, batch size, and scoring rules follow each evaluation's standard protocol. Public reference models (e.g.\ Qwen3-VL and InternVL-family models) are evaluated separately using their model-specific published decoding presets.

\begin{table}[h]
    \centering
    \small
    \begin{tabular}{ll}
        \toprule
        \multicolumn{2}{l}{\emph{Decoding}} \\
        \midrule
        Sampling strategy & Greedy \\
        Temperature & 0.0 \\
        Top-$p$ & n/a \\
        Top-$k$ & n/a \\
        Repetition penalty & none \\
        Maximum new tokens & 2{,}048 \\
        Stop sequences & none \\
        \midrule
        \multicolumn{2}{l}{\emph{Prompting}} \\
        \midrule
        System prompt & none \\
        Prompt template & Benchmark-specific MCQ or generative template \\
        Few-shot examples & 0 \\
        \midrule
        \multicolumn{2}{l}{\emph{Inference runtime}} \\
        \midrule
        Image preprocessing & Matches training \\
        Precision & bf16 autocast \\
        Batch size (inference) & 64 \\
        Framework & \texttt{datvlmeval} native checkpoint inference \\
        Decoding seed & n/a \\
        \midrule
        \multicolumn{2}{l}{\emph{Scoring}} \\
        \midrule
        Grounding (RefCOCO family) & center accuracy, IoU, Recall@$\{0.3, 0.5, 0.7\}$ \\
        Counting (CountBench) & exact / within-1 / within-3 \\
        Multi-image (BLINK) & per-category accuracy \\
        Other \datbench{} axes & LM-as-judge \\
        \bottomrule
    \end{tabular}
    \caption{Inference and evaluation settings for the DatologyAI-curated and MAmmoTH-VL \base checkpoints. Public references are evaluated with their own model-specific decoding presets but the same benchmark-side evaluation protocols.}
    \label{tab:inference-settings}
\end{table}

\subsection{Per-benchmark scores across all models and scales}
\label{app:results-aggregate-tables}

This subsection gives per-benchmark mean $\pm$ std scores for every internally trained model (MAmmoTH-VL \base and DatologyAI-curated at 1B, 2B, 4B) alongside every external reference model, on (i) the full 20-eval public suite (Table~\ref{tab:appendix-all-public}) and (ii) the 9-capability \datbench{} suite (Table~\ref{tab:appendix-datbench}). The aggregated figures in the main text (\S\ref{sec:main-results}, \S\ref{sec:robustness}) take the Avg column or per-capability columns of these tables. The seed-variance aggregate quoted in \S\ref{sec:robustness} (mean per-capability seed std drops from $2.47$ to $0.82$, $-67\%$, underlying Figure~\ref{fig:context-length-variance}) is the mean of the per-capability Std cells of Table~\ref{tab:appendix-datbench} at 2B.

\subsubsection{All public benchmarks (20 evals)}
\label{app:results-public}

Table~\ref{tab:appendix-all-public} covers the 20-eval public suite shown in the hero Pareto (Figure~\ref{fig:hero}); this is the union of the 11 IID benchmarks summarized in Figure~\ref{fig:general-capability-bars} and the 9 OOD benchmarks summarized in Figure~\ref{fig:ood-capability-bars} (including BLINK and the ecommerce brand-ID component). Evals are rows, models are columns; MAmmoTH-VL/Datology pairs are the matched-compute \base / DatologyAI-curated runs at each scale.

\begin{table}[h]
    \centering
    \scriptsize
    \resizebox{\linewidth}{!}{
    \begin{tabular}{lccccccccccccc}
        \toprule
        & \multicolumn{6}{c}{\textit{Ours} (mean $\pm$ std across seeds)} & \multicolumn{7}{c}{\textit{External} (single run)} \\
        \cmidrule(lr){2-7} \cmidrule(lr){8-14}
        Eval & MAmmoTH-VL 1B & Datology 1B & MAmmoTH-VL 2B & Datology 2B & MAmmoTH-VL 4B & Datology 4B & InternVL3-2B & InternVL3.5-2B & InternVL3.5-4B & Qwen3-VL-2B & Qwen3-VL-4B & Qwen3.5-2B & Qwen3.5-4B \\
        \midrule
        RefCOCO       & 44.8 {\scriptsize $\pm$ 0.7}  & 82.9 {\scriptsize $\pm$ 0.2} & 44.3 {\scriptsize $\pm$ 11.5} & 85.1 {\scriptsize $\pm$ 0.7} & 33.2 {\scriptsize $\pm$ 10.4} & 89.2 {\scriptsize $\pm$ 0.8} & 49.4 & 22.1 & 78.0 & 88.5 & 89.5 & 89.1 & 87.6 \\
        RefCOCOG      & 34.7 {\scriptsize $\pm$ 4.4}  & 77.0 {\scriptsize $\pm$ 0.6} & 38.8 {\scriptsize $\pm$ 11.3} & 79.5 {\scriptsize $\pm$ 0.6} & 29.4 {\scriptsize $\pm$ 9.5}  & 83.3 {\scriptsize $\pm$ 0.4} & 50.2 & 26.6 & 70.9 & 82.8 & 85.5 & 83.7 & 83.4 \\
        RefCOCO+      & 39.9 {\scriptsize $\pm$ 0.7}  & 74.5 {\scriptsize $\pm$ 0.0} & 40.0 {\scriptsize $\pm$ 10.8} & 76.2 {\scriptsize $\pm$ 1.2} & 30.2 {\scriptsize $\pm$ 10.7} & 82.7 {\scriptsize $\pm$ 1.7} & 45.3 & 21.0 & 72.3 & 82.9 & 87.0 & 83.6 & 83.6 \\
        PixMo Points  & 2.4 {\scriptsize $\pm$ 0.1}   & 15.4 {\scriptsize $\pm$ 0.6} & 1.5 {\scriptsize $\pm$ 0.3}   & 19.5 {\scriptsize $\pm$ 1.2} & 2.8 {\scriptsize $\pm$ 0.9}   & 23.2 {\scriptsize $\pm$ 1.9} & 3.6  & 10.9 & 13.4 & 9.0  & 9.8  & 10.2 & 14.8 \\
        MMBench       & 61.5 {\scriptsize $\pm$ 0.5}  & 63.3 {\scriptsize $\pm$ 1.7} & 69.7 {\scriptsize $\pm$ 1.2}  & 70.3 {\scriptsize $\pm$ 0.5} & 74.4 {\scriptsize $\pm$ 1.2}  & 76.7 {\scriptsize $\pm$ 0.3} & 72.2 & 75.7 & 68.6 & 75.7 & 83.6 & 78.0 & 81.8 \\
        RealWorldQA  & 43.3 {\scriptsize $\pm$ 1.5}  & 53.7 {\scriptsize $\pm$ 0.8} & 47.2 {\scriptsize $\pm$ 1.6}  & 57.9 {\scriptsize $\pm$ 0.4} & 53.3 {\scriptsize $\pm$ 0.7}  & 61.8 {\scriptsize $\pm$ 1.4} & 54.4 & 51.5 & 57.9 & 55.4 & 64.7 & 56.7 & 58.3 \\
        TextVQA       & 65.1 {\scriptsize $\pm$ 0.5}  & 67.7 {\scriptsize $\pm$ 1.2} & 68.7 {\scriptsize $\pm$ 0.3}  & 73.4 {\scriptsize $\pm$ 0.4} & 72.5 {\scriptsize $\pm$ 1.7}  & 75.0 {\scriptsize $\pm$ 0.5} & 72.7 & 67.3 & 70.5 & 74.6 & 76.2 & 68.6 & 74.3 \\
        OCRBench      & 61.3 {\scriptsize $\pm$ 1.0}  & 65.8 {\scriptsize $\pm$ 0.2} & 66.9 {\scriptsize $\pm$ 0.7}  & 70.6 {\scriptsize $\pm$ 0.9} & 68.0 {\scriptsize $\pm$ 1.4}  & 73.7 {\scriptsize $\pm$ 0.4} & 84.1 & 83.4 & 82.7 & 84.4 & 86.2 & 86.3 & 89.3 \\
        DocVQA        & 72.9 {\scriptsize $\pm$ 1.2}  & 80.8 {\scriptsize $\pm$ 0.0} & 80.5 {\scriptsize $\pm$ 0.4}  & 85.1 {\scriptsize $\pm$ 0.2} & 83.7 {\scriptsize $\pm$ 0.4}  & 88.9 {\scriptsize $\pm$ 0.2} & 84.1 & 84.9 & 89.0 & 90.9 & 93.2 & 87.7 & 92.3 \\
        DetailCaps    & 63.2 {\scriptsize $\pm$ 0.1}  & 62.9 {\scriptsize $\pm$ 0.1} & 63.1 {\scriptsize $\pm$ 0.1}  & 63.5 {\scriptsize $\pm$ 0.3} & 62.8 {\scriptsize $\pm$ 0.1}  & 64.2 {\scriptsize $\pm$ 0.0} & 64.2 & 64.1 & 63.8 & 63.9 & 63.0 & 61.8 & 57.2 \\
        CAPability    & 59.2 {\scriptsize $\pm$ 0.2}  & 61.0 {\scriptsize $\pm$ 0.6} & 60.9 {\scriptsize $\pm$ 0.1}  & 61.7 {\scriptsize $\pm$ 0.3} & 62.2 {\scriptsize $\pm$ 0.4}  & 64.5 {\scriptsize $\pm$ 1.5} & 61.8 & 61.5 & 61.0 & 71.1 & 74.2 & 74.5 & 77.0 \\
        CVBench-2D    & 30.6 {\scriptsize $\pm$ 2.1}  & 59.8 {\scriptsize $\pm$ 0.6} & 47.2 {\scriptsize $\pm$ 2.4}  & 65.2 {\scriptsize $\pm$ 1.8} & 53.5 {\scriptsize $\pm$ 1.9}  & 71.6 {\scriptsize $\pm$ 0.0} & 60.8 & 63.6 & 62.9 & 60.0 & 72.1 & 62.9 & 72.4 \\
        CVBench-3D    & 44.4 {\scriptsize $\pm$ 11.2} & 75.1 {\scriptsize $\pm$ 4.8} & 55.8 {\scriptsize $\pm$ 3.1}  & 76.3 {\scriptsize $\pm$ 2.7} & 62.2 {\scriptsize $\pm$ 4.8}  & 78.3 {\scriptsize $\pm$ 1.0} & 69.4 & 66.9 & 78.8 & 79.2 & 89.1 & 81.0 & 84.1 \\
        3DSRBench     & 32.5 {\scriptsize $\pm$ 0.1}  & 42.8 {\scriptsize $\pm$ 0.8} & 32.9 {\scriptsize $\pm$ 3.0}  & 45.9 {\scriptsize $\pm$ 0.3} & 42.0 {\scriptsize $\pm$ 1.4}  & 48.0 {\scriptsize $\pm$ 0.0} & 38.8 & 43.0 & 32.8 & 43.2 & 46.5 & 47.0 & 45.0 \\
        CountBench    & 83.5 {\scriptsize $\pm$ 0.2}  & 88.8 {\scriptsize $\pm$ 0.4} & 86.1 {\scriptsize $\pm$ 0.4}  & 89.9 {\scriptsize $\pm$ 0.7} & 82.9 {\scriptsize $\pm$ 1.0}  & 90.2 {\scriptsize $\pm$ 0.6} & 66.2 & 71.7 & 80.2 & 80.4 & 81.3 & 85.3 & 89.8 \\
        AI2D          & 61.7 {\scriptsize $\pm$ 0.3}  & 65.5 {\scriptsize $\pm$ 0.3} & 70.5 {\scriptsize $\pm$ 0.7}  & 73.7 {\scriptsize $\pm$ 0.2} & 74.5 {\scriptsize $\pm$ 0.1}  & 77.6 {\scriptsize $\pm$ 0.4} & 75.6 & 74.7 & 78.7 & 74.2 & 81.1 & 77.4 & 76.5 \\
        ChartQA       & 72.4 {\scriptsize $\pm$ 0.6}  & 78.7 {\scriptsize $\pm$ 0.2} & 78.9 {\scriptsize $\pm$ 0.1}  & 81.7 {\scriptsize $\pm$ 0.1} & 81.4 {\scriptsize $\pm$ 0.6}  & 84.0 {\scriptsize $\pm$ 0.5} & 78.4 & 74.8 & 84.0 & 76.8 & 82.8 & 83.1 & 82.3 \\
        MathVista     & 40.1 {\scriptsize $\pm$ 0.0}  & 45.6 {\scriptsize $\pm$ 0.2} & 49.8 {\scriptsize $\pm$ 0.8}  & 53.6 {\scriptsize $\pm$ 0.9} & 52.8 {\scriptsize $\pm$ 2.6}  & 57.4 {\scriptsize $\pm$ 0.6} & 60.9 & 61.4 & 67.0 & 59.7 & 72.0 & 71.4 & 64.1 \\
        Brand ID      & 23.4 {\scriptsize $\pm$ 7.8}  & 34.0 {\scriptsize $\pm$ 0.4} & 31.5 {\scriptsize $\pm$ 4.0}  & 36.7 {\scriptsize $\pm$ 0.2} & 28.2 {\scriptsize $\pm$ 3.3}  & 37.1 {\scriptsize $\pm$ 0.0} & 35.4 & 35.2 & 36.3 & 36.7 & 31.6 & 37.3 & 38.5 \\
        BLINK         & 40.3 {\scriptsize $\pm$ 0.7}  & 42.5 {\scriptsize $\pm$ 0.3} & 41.7 {\scriptsize $\pm$ 0.6}  & 44.8 {\scriptsize $\pm$ 0.7} & 44.3 {\scriptsize $\pm$ 1.5}  & 45.9 {\scriptsize $\pm$ 0.7} & 44.2 & 52.6 & 57.8 & 55.7 & 66.6 & 62.0 & 63.5 \\
        \midrule
        \textbf{Avg}  & \textbf{48.9} {\scriptsize $\pm$ 0.7} & \textbf{61.9} {\scriptsize $\pm$ 0.3} & \textbf{53.8} {\scriptsize $\pm$ 1.4} & \textbf{65.5} {\scriptsize $\pm$ 0.0} & \textbf{54.7} {\scriptsize $\pm$ 1.4} & \textbf{68.7} {\scriptsize $\pm$ 0.3} & \textbf{58.6} & \textbf{55.6} & \textbf{65.3} & \textbf{67.3} & \textbf{71.8} & \textbf{69.4} & \textbf{70.8} \\
        \bottomrule
    \end{tabular}
    }
    \caption{Per-benchmark scores on the full 20-eval public suite (the union of the 11 IID benchmarks in Figure~\ref{fig:general-capability-bars} and the 9 OOD benchmarks in Figure~\ref{fig:ood-capability-bars}; same set used by the hero Pareto in Figure~\ref{fig:hero}). \emph{MAmmoTH-VL} columns are the matched-compute MAmmoTH-VL-12M single-image \base; \emph{Datology} columns are the DatologyAI-curated mixture under the same recipe and compute. Internal cells show mean $\pm$ std across training seeds; external models are evaluated as released. The bottom \textbf{Avg} row reports each model's across-eval mean; the $\pm$ on Datology Avg cells is the std of the per-seed across-eval means (i.e., how much the suite-level average varies seed-to-seed), distinct from the mean per-cell std reported as the \datbench{} reliability headline in \S\ref{sec:robustness}.}
    \label{tab:appendix-all-public}
\end{table}

\subsubsection{\datbench{} (9 capabilities)}
\label{app:results-datbench}

Table~\ref{tab:appendix-datbench} gives the per-capability \datbench{} breakdown for every internal model and external reference. The 2B row pair underlies Figure~\ref{fig:main-capability-bars}; the 1B/2B/4B Datology rows underlie the scaling figure (Figure~\ref{fig:scale-across}).

\begin{table}[h]
    \centering
    \scriptsize
    \resizebox{\linewidth}{!}{
    \begin{tabular}{lccccccccccccc}
        \toprule
        & \multicolumn{6}{c}{\textit{Ours} (mean $\pm$ std across seeds)} & \multicolumn{7}{c}{\textit{External} (single run)} \\
        \cmidrule(lr){2-7} \cmidrule(lr){8-14}
        Capability & MAmmoTH-VL 1B & Datology 1B & MAmmoTH-VL 2B & Datology 2B & MAmmoTH-VL 4B & Datology 4B & InternVL3-2B & InternVL3.5-2B & InternVL3.5-4B & Qwen3-VL-2B & Qwen3-VL-4B & Qwen3.5-2B & Qwen3.5-4B \\
        \midrule
        Chart       & 32.6 {\scriptsize $\pm$ 1.1} & 44.9 {\scriptsize $\pm$ 0.4} & 39.8 {\scriptsize $\pm$ 0.5}  & 51.7 {\scriptsize $\pm$ 0.9} & 45.2 {\scriptsize $\pm$ 1.3}  & 57.8 {\scriptsize $\pm$ 0.1} & 49.1 & 55.6 & 62.9 & 56.5 & 64.6 & 65.8 & 69.0 \\
        Counting    & 77.8 {\scriptsize $\pm$ 0.8} & 78.4 {\scriptsize $\pm$ 0.5} & 80.0 {\scriptsize $\pm$ 1.4}  & 82.5 {\scriptsize $\pm$ 0.8} & 81.6 {\scriptsize $\pm$ 0.3}  & 84.4 {\scriptsize $\pm$ 0.9} & 82.9 & 62.8 & 82.6 & 84.3 & 89.1 & 83.5 & 80.3 \\
        Document    & 35.5 {\scriptsize $\pm$ 0.4} & 44.2 {\scriptsize $\pm$ 0.6} & 39.5 {\scriptsize $\pm$ 0.9}  & 49.1 {\scriptsize $\pm$ 0.1} & 43.2 {\scriptsize $\pm$ 0.7}  & 52.2 {\scriptsize $\pm$ 0.1} & 50.8 & 52.2 & 47.3 & 60.1 & 63.2 & 56.8 & 51.7 \\
        General     & 48.3 {\scriptsize $\pm$ 0.4} & 51.1 {\scriptsize $\pm$ 0.0} & 55.9 {\scriptsize $\pm$ 0.5}  & 57.5 {\scriptsize $\pm$ 0.1} & 60.7 {\scriptsize $\pm$ 0.4}  & 62.8 {\scriptsize $\pm$ 0.4} & 60.5 & 55.7 & 61.5 & 65.2 & 75.9 & 65.6 & 67.8 \\
        Grounding   & 31.0 {\scriptsize $\pm$ 0.9} & 72.3 {\scriptsize $\pm$ 1.1} & 18.3 {\scriptsize $\pm$ 14.2} & 75.4 {\scriptsize $\pm$ 1.7} & 29.1 {\scriptsize $\pm$ 11.7} & 81.7 {\scriptsize $\pm$ 0.9} & 51.0 & 30.6 & 79.0 & 82.6 & 87.5 & 81.2 & 84.2 \\
        Math        & 13.1 {\scriptsize $\pm$ 0.5} & 14.1 {\scriptsize $\pm$ 0.1} & 16.8 {\scriptsize $\pm$ 0.2}  & 19.1 {\scriptsize $\pm$ 0.3} & 18.2 {\scriptsize $\pm$ 0.8}  & 23.1 {\scriptsize $\pm$ 0.3} & 16.9 & 24.9 & 25.0 & 22.8 & 33.3 & 33.6 & 24.7 \\
        Scene       & 57.3 {\scriptsize $\pm$ 1.0} & 61.4 {\scriptsize $\pm$ 0.4} & 61.1 {\scriptsize $\pm$ 0.2}  & 67.8 {\scriptsize $\pm$ 0.5} & 63.5 {\scriptsize $\pm$ 2.2}  & 72.0 {\scriptsize $\pm$ 0.5} & 67.4 & 60.5 & 64.9 & 79.9 & 83.2 & 58.1 & 40.0 \\
        Spatial     & 36.2 {\scriptsize $\pm$ 2.9} & 42.5 {\scriptsize $\pm$ 0.7} & 41.0 {\scriptsize $\pm$ 2.6}  & 45.3 {\scriptsize $\pm$ 2.6} & 48.3 {\scriptsize $\pm$ 1.2}  & 55.2 {\scriptsize $\pm$ 1.1} & 38.6 & 30.1 & 54.0 & 37.7 & 39.1 & 31.1 & 29.1 \\
        Table       & 35.5 {\scriptsize $\pm$ 1.0} & 36.8 {\scriptsize $\pm$ 0.0} & 36.2 {\scriptsize $\pm$ 1.8}  & 42.4 {\scriptsize $\pm$ 0.5} & 44.1 {\scriptsize $\pm$ 1.6}  & 49.7 {\scriptsize $\pm$ 0.4} & 37.6 & 51.1 & 57.9 & 49.5 & 63.3 & 50.9 & 24.5 \\
        \midrule
        \textbf{Avg} & \textbf{40.8} {\scriptsize $\pm$ 0.5} & \textbf{49.5} {\scriptsize $\pm$ 0.3} & \textbf{43.2} {\scriptsize $\pm$ 1.7} & \textbf{54.5} {\scriptsize $\pm$ 0.5} & \textbf{48.2} {\scriptsize $\pm$ 1.4} & \textbf{59.9} {\scriptsize $\pm$ 0.0} & \textbf{50.5} & \textbf{47.1} & \textbf{59.4} & \textbf{59.8} & \textbf{66.6} & \textbf{58.5} & \textbf{52.4} \\
        \bottomrule
    \end{tabular}
    }
    \caption{Per-capability scores on the 9-capability \datbench{} suite. \emph{MAmmoTH-VL} columns are the matched-compute MAmmoTH-VL-12M single-image \base; \emph{Datology} columns are the DatologyAI-curated mixture. Internal cells show mean $\pm$ std across training seeds; external models are evaluated as released. The 2B MAmmoTH-VL/Datology column pair underlies the per-capability bars in Figure~\ref{fig:main-capability-bars}; the 1B/2B/4B Datology columns underlie the scaling story in Figure~\ref{fig:scale-across}. Capability column abbreviations: Chart $=$ Chart Understanding, Document $=$ Document Understanding, Scene $=$ Scene OCR, Table $=$ Diagrams \& Tables, Math $=$ Math \& Logic, Spatial $=$ Spatial Reasoning. The Avg column $\pm$ on Datology rows is the std of per-seed across-capability means; the mean per-capability std (the $-67\%$ headline in \S\ref{sec:robustness}) is the row-wise mean of the Std cells.}
    \label{tab:appendix-datbench}
\end{table}

\subsection{Mean response tokens per benchmark}
\label{app:results-tokens}

Table~\ref{tab:appendix-tokens-public} reports the mean number of generated response tokens per benchmark for every model in the comparison set, on the same 20-eval public suite as Table~\ref{tab:appendix-all-public}. The Avg column is the across-eval mean and is what drives the response-FLOPs axis in the inference-efficiency Pareto (Figure~\ref{fig:inference-efficiency-pareto}, \S\ref{sec:inference-efficiency}): response-FLOPs $= 2 \cdot \text{active params} \cdot \text{Avg tokens}$.

\begin{table}[h]
    \centering
    \scriptsize
    \resizebox{\linewidth}{!}{
    \begin{tabular}{lccccccccccccc}
        \toprule
        & \multicolumn{6}{c}{\textit{Ours}} & \multicolumn{7}{c}{\textit{External}} \\
        \cmidrule(lr){2-7} \cmidrule(lr){8-14}
        Eval & MAmmoTH-VL 1B & Datology 1B & MAmmoTH-VL 2B & Datology 2B & MAmmoTH-VL 4B & Datology 4B & InternVL3-2B & InternVL3.5-2B & InternVL3.5-4B & Qwen3-VL-2B & Qwen3-VL-4B & Qwen3.5-2B & Qwen3.5-4B \\
        \midrule
        RefCOCO       & 24.1  & 19.6  & 21.5  & 22.1  & 21.2  & 22.5  & 29.5  & 150.6 & 23.1  & 22.5  & 22.2  & 23.7  & 744.3 \\
        RefCOCOG      & 23.6  & 19.6  & 21.2  & 21.9  & 21.7  & 22.6  & 29.9  & 120.7 & 23.1  & 22.5  & 22.3  & 23.8  & 830.8 \\
        RefCOCO+      & 24.8  & 19.6  & 21.7  & 22.1  & 21.5  & 22.5  & 32.3  & 147.5 & 23.3  & 22.6  & 22.1  & 23.7  & 783.5 \\
        PixMo Points  & 9.8   & 11.7  & 9.1   & 11.7  & 29.3  & 11.5  & 35.0  & 132.1 & 34.5  & 14.7  & 44.3  & 79.1  & 1129.5 \\
        MMBench       & 1.1   & 1.0   & 1.0   & 1.0   & 1.2   & 1.0   & 11.6  & 132.6 & 108.5 & 100.2 & 155.6 & 171.3 & 2684.6 \\
        RealWorldQA  & 1.2   & 1.1   & 1.2   & 1.1   & 1.2   & 1.3   & 5.3   & 70.6  & 30.2  & 6.2   & 57.4  & 86.3  & 2167.0 \\
        TextVQA       & 7.2   & 6.0   & 5.3   & 4.1   & 8.3   & 4.9   & 3.2   & 22.3  & 42.4  & 6.7   & 5.0   & 35.8  & 702.0 \\
        OCRBench      & 6.5   & 9.9   & 7.1   & 8.5   & 7.0   & 12.4  & 9.9   & 25.5  & 28.5  & 37.0  & 34.7  & 36.5  & 549.4 \\
        DocVQA        & 5.6   & 5.9   & 5.2   & 5.4   & 5.5   & 5.8   & 6.1   & 35.5  & 6.3   & 6.5   & 5.7   & 38.5  & 360.1 \\
        DetailCaps    & 456.3 & 138.3 & 474.9 & 298.9 & 468.7 & 325.4 & 173.7 & 178.8 & 155.1 & 282.6 & 358.0 & 408.6 & 1229.0 \\
        CAPability    & 517.6 & 396.7 & 526.7 & 425.5 & 520.8 & 202.2 & 139.7 & 210.8 & 159.1 & 289.8 & 373.7 & 435.7 & 1271.3 \\
        CVBench-2D    & 1.5   & 1.0   & 1.0   & 1.0   & 1.1   & 1.0   & 3.6   & 101.1 & 28.5  & 9.1   & 103.4 & 123.1 & 2832.5 \\
        CVBench-3D    & 1.4   & 1.0   & 1.0   & 1.0   & 1.0   & 1.0   & 3.1   & 188.8 & 113.5 & 5.0   & 11.9  & 103.8 & 1564.0 \\
        3DSRBench     & 1.1   & 1.0   & 1.0   & 1.0   & 20.0  & 1.0   & 4.1   & 147.2 & 51.4  & 19.5  & 40.1  & 179.6 & 2921.6 \\
        CountBench    & 1.8   & 1.2   & 1.2   & 1.4   & 1.5   & 1.4   & 2.8   & 92.2  & 86.5  & 50.8  & 8.7   & 42.5  & 562.8 \\
        AI2D          & 1.0   & 1.0   & 1.0   & 1.0   & 1.0   & 1.0   & 6.0   & 157.9 & 229.5 & 302.3 & 88.3  & 338.0 & 1216.1 \\
        ChartQA       & 4.0   & 3.5   & 3.5   & 4.3   & 3.5   & 3.6   & 3.7   & 18.8  & 18.4  & 7.6   & 11.8  & 78.1  & 658.3 \\
        MathVista     & 14.5  & 2.8   & 5.8   & 2.6   & 37.7  & 2.1   & 77.5  & 278.9 & 255.0 & 446.0 & 458.0 & 524.4 & 1282.9 \\
        Brand ID      & 11.2  & 6.2   & 5.7   & 5.4   & 16.1  & 5.9   & 14.1  & 114.0 & 76.1  & 84.8  & 68.5  & 162.8 & 778.0 \\
        BLINK         & 31.1  & 1.0   & 1.5   & 1.0   & 3.5   & 1.1   & 8.6   & 244.8 & 214.6 & 295.0 & 285.3 & 472.2 & 1410.8 \\
        \midrule
        \textbf{Avg}  & \textbf{57.3} & \textbf{32.4} & \textbf{55.8} & \textbf{42.1} & \textbf{59.6} & \textbf{32.5} & \textbf{30.0} & \textbf{128.5} & \textbf{85.4} & \textbf{101.6} & \textbf{108.9} & \textbf{169.4} & \textbf{1283.9} \\
        \bottomrule
    \end{tabular}
    }
    \caption{Mean generated-response token count per benchmark on the same 20-eval public suite as Table~\ref{tab:appendix-all-public}, averaged across seeds for our checkpoints and across released runs for externals. The bottom \textbf{Avg} row is the across-eval mean used as the response-tokens factor in the response-FLOPs proxy ($2 \cdot \text{active params} \cdot \text{Avg tokens}$) for the inference-efficiency Pareto in Figure~\ref{fig:inference-efficiency-pareto} (\S\ref{sec:inference-efficiency}). The two captioning evals (DetailCaps, CAPability) and the post-trained references (notably Qwen3.5-4B) account for most of the variation in the Avg row.}
    \label{tab:appendix-tokens-public}
\end{table}

\subsection{FLOPs methodology}
\label{app:results-flops}

We report two FLOPs metrics in the paper: \emph{training FLOPs} on the hero Pareto's $x$-axis (Figure~\ref{fig:hero}) and \emph{response FLOPs} on the inference-efficiency Pareto's $x$-axis (Figure~\ref{fig:inference-efficiency-pareto}, \S\ref{sec:inference-efficiency}). Both follow the standard transformer cost approximations:
\begin{align*}
F_{\text{train}}     &\;=\; 6 \cdot N \cdot D, \\
F_{\text{response}}  &\;=\; 2 \cdot N \cdot T,
\end{align*}
where $N$ is the model's active parameter count, $D$ is its number of vision--language (VL) training tokens, and $T$ is its mean generated-response token count averaged across the 20 evals in Table~\ref{tab:appendix-tokens-public}. The $6N$ approximation is the standard transformer-training cost (forward $+$ backward $+$ activation recompute); the $2N$ inference cost counts the decode-only forward pass per generated token.

We restrict $D$ to \emph{VL-stage} tokens so the comparison is apples-to-apples across models with different language-only pretraining histories: language-only LLM pretraining is folded into $N$ (it is part of the backbone) and is not double-counted in $D$. Under this convention our matched-compute MAmmoTH-VL \base and DatologyAI-curated runs both use $D = 25$B VL tokens at every scale.

All MAmmoTH-VL/Datology checkpoints share architecture: a Qwen3 LLM backbone (unfrozen during VL training), a 400M-parameter SigLIP-Large ViT, and a 2-layer MLP projector. Per-model $N$, $D$, $T$, and the resulting $F_{\text{train}}$ and $F_{\text{response}}$ are summarized in Table~\ref{tab:flops-methodology}; sources for $N$ and $D$ on external references are model cards / release notes.

As a worked example for the \curated 2B: $N = 2.10\times10^9$ (1.7B Qwen3 LLM + 0.4B SigLIP ViT), $D = 25\times10^9$ VL tokens, $T = 42.1$ (the 20-eval mean from Table~\ref{tab:appendix-tokens-public}), giving $F_{\text{train}} = 6 \cdot 2.10\times10^9 \cdot 25\times10^9 = 3.15\times10^{20}$ FLOPs and $F_{\text{response}} = 2 \cdot 2.10\times10^9 \cdot 42.1 = 1.77\times10^{11}$ FLOPs.

\begin{table}[h]
    \centering
    \footnotesize
    \resizebox{\linewidth}{!}{
    \begin{tabular}{lcrrrrr}
        \toprule
        Model & LLM backbone & N & D (VL tokens) & T (resp tokens) & $F_{\text{train}}$ & $F_{\text{response}}$ \\
        \midrule
        MAmmoTH-VL \Base 1B & Qwen3-0.6B & 1.00B & 25B & 57.3 & 1.50$\times$10$^{20}$ & 1.15$\times$10$^{11}$ \\
        Datology Curation 1B & Qwen3-0.6B & 1.00B & 25B & 32.4 & 1.50$\times$10$^{20}$ & 6.48$\times$10$^{10}$ \\
        MAmmoTH-VL \Base 2B & Qwen3-1.7B & 2.10B & 25B & 55.8 & 3.15$\times$10$^{20}$ & 2.34$\times$10$^{11}$ \\
        Datology Curation 2B & Qwen3-1.7B & 2.10B & 25B & 42.1 & 3.15$\times$10$^{20}$ & 1.77$\times$10$^{11}$ \\
        MAmmoTH-VL \Base 4B & Qwen3-4B & 4.40B & 25B & 59.6 & 6.60$\times$10$^{20}$ & 5.24$\times$10$^{11}$ \\
        Datology Curation 4B & Qwen3-4B & 4.40B & 25B & 32.5 & 6.60$\times$10$^{20}$ & 2.86$\times$10$^{11}$ \\
        InternVL3-2B (MPO) & Qwen2.5-1.5B & 2.09B & 244B & 30.0 & 3.06$\times$10$^{21}$ & 1.25$\times$10$^{11}$ \\
        InternVL3.5-2B (CascadeRL) & Qwen3-1.7B & 2.30B & 381B & 128.5 & 5.26$\times$10$^{21}$ & 5.91$\times$10$^{11}$ \\
        InternVL3.5-4B (CascadeRL) & Qwen3-4B & 4.30B & 381B & 85.4 & 9.83$\times$10$^{21}$ & 7.34$\times$10$^{11}$ \\
        Qwen3-VL-2B & Qwen3-1.7B & 2.10B & 2.17T & 101.6 & 2.73$\times$10$^{22}$ & 4.27$\times$10$^{11}$ \\
        Qwen3-VL-4B & Qwen3-4B & 4.30B & 2.17T & 108.9 & 5.60$\times$10$^{22}$ & 9.36$\times$10$^{11}$ \\
        Qwen3.5-2B & Qwen3-1.7B & 2.00B & 4.00T & 169.4 & 4.80$\times$10$^{22}$ & 6.77$\times$10$^{11}$ \\
        Qwen3.5-4B & Qwen3-4B & 4.00B & 4.00T & 1283.9 & 9.60$\times$10$^{22}$ & 1.03$\times$10$^{13}$ \\
        \bottomrule
    \end{tabular}
    }
    \caption{Per-model active parameters ($N$), VL training tokens ($D$), 20-eval mean response-token count ($T$, from Table~\ref{tab:appendix-tokens-public}), and the resulting training FLOPs $F_{\text{train}} = 6ND$ and response FLOPs $F_{\text{response}} = 2NT$ used in Figure~\ref{fig:hero} and Figure~\ref{fig:inference-efficiency-pareto} respectively. Sources for external $N$ and $D$ are the corresponding model cards / release notes.}
    \label{tab:flops-methodology}
\end{table}

\subsection{Context-length sweep}
\label{app:results-context}

Tables~\ref{tab:appendix-context-public} and~\ref{tab:appendix-context-datbench} give the per-benchmark / per-capability breakdown of the 4k / 8k / 16k context-length sweep at 2B, for both MAmmoTH-VL \base and DatologyAI-curated mixtures. The Avg row aggregates underlie the curation-gain figure (Figure~\ref{fig:context-length}) and the seed-variance-vs-context-length figure (Figure~\ref{fig:context-length-variance}) in \S\ref{sec:robustness}.

\begin{table}[h]
    \centering
    \scriptsize
    \resizebox{\linewidth}{!}{
    \begin{tabular}{lcccccc}
        \toprule
        & \multicolumn{2}{c}{4k context} & \multicolumn{2}{c}{8k context} & \multicolumn{2}{c}{16k context} \\
        \cmidrule(lr){2-3} \cmidrule(lr){4-5} \cmidrule(lr){6-7}
        Eval & MAmmoTH-VL 2B & Datology 2B & MAmmoTH-VL 2B & Datology 2B & MAmmoTH-VL 2B & Datology 2B \\
        \midrule
        RefCOCO       & 44.3 {\scriptsize $\pm$ 11.5} & 85.1 {\scriptsize $\pm$ 0.7} & 29.4 {\scriptsize $\pm$ 18.5} & 81.9 {\scriptsize $\pm$ 1.9} & 13.3 {\scriptsize $\pm$ 13.9} & 81.2 {\scriptsize $\pm$ 2.5} \\
        RefCOCOG      & 38.8 {\scriptsize $\pm$ 11.3} & 79.5 {\scriptsize $\pm$ 0.6} & 28.9 {\scriptsize $\pm$ 15.8} & 75.2 {\scriptsize $\pm$ 0.0} & 14.3 {\scriptsize $\pm$ 15.2} & 74.4 {\scriptsize $\pm$ 2.3} \\
        RefCOCO+      & 40.0 {\scriptsize $\pm$ 10.8} & 76.2 {\scriptsize $\pm$ 1.2} & 26.3 {\scriptsize $\pm$ 15.4} & 73.6 {\scriptsize $\pm$ 1.4} & 12.3 {\scriptsize $\pm$ 13.2} & 71.8 {\scriptsize $\pm$ 2.5} \\
        PixMo Points  & 1.5 {\scriptsize $\pm$ 0.3}   & 19.5 {\scriptsize $\pm$ 1.2} & 0.9 {\scriptsize $\pm$ 0.9}   & 2.3 {\scriptsize $\pm$ 1.2}  & 0.4 {\scriptsize $\pm$ 0.0}   & 1.4 {\scriptsize $\pm$ 0.3} \\
        MMBench       & 69.7 {\scriptsize $\pm$ 1.2}  & 70.3 {\scriptsize $\pm$ 0.5} & 69.1 {\scriptsize $\pm$ 0.5}  & 69.9 {\scriptsize $\pm$ 0.8} & 68.8 {\scriptsize $\pm$ 0.2}  & 69.3 {\scriptsize $\pm$ 0.2} \\
        RealWorldQA  & 47.2 {\scriptsize $\pm$ 1.6}  & 57.9 {\scriptsize $\pm$ 0.4} & 46.7 {\scriptsize $\pm$ 1.1}  & 54.5 {\scriptsize $\pm$ 0.0} & 48.4 {\scriptsize $\pm$ 1.6}  & 51.7 {\scriptsize $\pm$ 1.6} \\
        TextVQA       & 68.7 {\scriptsize $\pm$ 0.3}  & 73.4 {\scriptsize $\pm$ 0.4} & 69.5 {\scriptsize $\pm$ 1.3}  & 70.0 {\scriptsize $\pm$ 0.5} & 69.0 {\scriptsize $\pm$ 1.1}  & 68.7 {\scriptsize $\pm$ 1.4} \\
        OCRBench      & 66.9 {\scriptsize $\pm$ 0.7}  & 70.6 {\scriptsize $\pm$ 0.9} & 64.8 {\scriptsize $\pm$ 1.9}  & 67.6 {\scriptsize $\pm$ 0.6} & 63.0 {\scriptsize $\pm$ 0.1}  & 64.3 {\scriptsize $\pm$ 0.4} \\
        DocVQA        & 80.5 {\scriptsize $\pm$ 0.4}  & 85.1 {\scriptsize $\pm$ 0.2} & 79.0 {\scriptsize $\pm$ 0.8}  & 81.0 {\scriptsize $\pm$ 0.6} & 77.4 {\scriptsize $\pm$ 0.8}  & 80.4 {\scriptsize $\pm$ 0.5} \\
        DetailCaps    & 63.1 {\scriptsize $\pm$ 0.1}  & 63.5 {\scriptsize $\pm$ 0.3} & 62.8 {\scriptsize $\pm$ 0.1}  & 62.9 {\scriptsize $\pm$ 0.2} & 62.8 {\scriptsize $\pm$ 0.1}  & 63.4 {\scriptsize $\pm$ 0.0} \\
        CAPability    & 60.9 {\scriptsize $\pm$ 0.1}  & 61.7 {\scriptsize $\pm$ 0.3} & 60.4 {\scriptsize $\pm$ 0.3}  & 60.5 {\scriptsize $\pm$ 0.5} & 60.0 {\scriptsize $\pm$ 0.3}  & 61.6 {\scriptsize $\pm$ 0.6} \\
        CVBench-2D    & 47.2 {\scriptsize $\pm$ 2.4}  & 65.2 {\scriptsize $\pm$ 1.8} & 39.7 {\scriptsize $\pm$ 7.2}  & 61.3 {\scriptsize $\pm$ 1.6} & 40.1 {\scriptsize $\pm$ 6.2}  & 56.2 {\scriptsize $\pm$ 1.1} \\
        CVBench-3D    & 55.8 {\scriptsize $\pm$ 3.1}  & 76.3 {\scriptsize $\pm$ 2.7} & 54.8 {\scriptsize $\pm$ 2.4}  & 60.8 {\scriptsize $\pm$ 9.7} & 46.1 {\scriptsize $\pm$ 3.6}  & 58.9 {\scriptsize $\pm$ 0.2} \\
        3DSRBench     & 32.9 {\scriptsize $\pm$ 3.0}  & 45.9 {\scriptsize $\pm$ 0.3} & 31.4 {\scriptsize $\pm$ 2.1}  & 42.8 {\scriptsize $\pm$ 0.0} & 29.6 {\scriptsize $\pm$ 4.4}  & 39.5 {\scriptsize $\pm$ 0.5} \\
        CountBench    & 86.1 {\scriptsize $\pm$ 0.4}  & 89.9 {\scriptsize $\pm$ 0.7} & 85.9 {\scriptsize $\pm$ 0.6}  & 85.8 {\scriptsize $\pm$ 0.1} & 83.9 {\scriptsize $\pm$ 0.3}  & 85.4 {\scriptsize $\pm$ 0.1} \\
        AI2D          & 70.5 {\scriptsize $\pm$ 0.7}  & 73.7 {\scriptsize $\pm$ 0.2} & 69.7 {\scriptsize $\pm$ 0.1}  & 70.6 {\scriptsize $\pm$ 0.0} & 68.2 {\scriptsize $\pm$ 0.5}  & 68.9 {\scriptsize $\pm$ 0.2} \\
        ChartQA       & 78.9 {\scriptsize $\pm$ 0.1}  & 81.7 {\scriptsize $\pm$ 0.1} & 78.4 {\scriptsize $\pm$ 0.4}  & 77.2 {\scriptsize $\pm$ 0.5} & 77.2 {\scriptsize $\pm$ 0.7}  & 77.4 {\scriptsize $\pm$ 0.0} \\
        MathVista     & 49.8 {\scriptsize $\pm$ 0.8}  & 53.6 {\scriptsize $\pm$ 0.9} & 49.6 {\scriptsize $\pm$ 2.3}  & 47.1 {\scriptsize $\pm$ 1.4} & 46.6 {\scriptsize $\pm$ 0.7}  & 46.4 {\scriptsize $\pm$ 1.6} \\
        Brand ID      & 31.5 {\scriptsize $\pm$ 4.0}  & 36.7 {\scriptsize $\pm$ 0.2} & 27.2 {\scriptsize $\pm$ 5.6}  & 33.7 {\scriptsize $\pm$ 1.9} & 30.8 {\scriptsize $\pm$ 5.5}  & 28.6 {\scriptsize $\pm$ 1.6} \\
        BLINK         & 41.7 {\scriptsize $\pm$ 0.6}  & 44.8 {\scriptsize $\pm$ 0.7} & 41.7 {\scriptsize $\pm$ 0.1}  & 43.3 {\scriptsize $\pm$ 0.1} & 42.2 {\scriptsize $\pm$ 1.2}  & 43.4 {\scriptsize $\pm$ 0.5} \\
        \midrule
        \textbf{Avg}  & \textbf{53.8} {\scriptsize $\pm$ 1.4} & \textbf{65.5} {\scriptsize $\pm$ 0.0} & \textbf{50.8} {\scriptsize $\pm$ 2.4} & \textbf{61.1} {\scriptsize $\pm$ 0.8} & \textbf{47.7} {\scriptsize $\pm$ 2.8} & \textbf{59.6} {\scriptsize $\pm$ 0.3} \\
        \bottomrule
    \end{tabular}
    }
    \caption{Per-benchmark scores on the 20-eval public suite at 2B across the 4k / 8k / 16k context-length sweep, \base (MAmmoTH-VL) vs.\ DatologyAI-curated. Cells are mean $\pm$ std across training seeds (Datology has 3 / 2 / 2 seeds at 4k / 8k / 16k respectively). The Avg row gives a $+11.7$ / $+10.3$ / $+11.9$pp curation gain at 4k / 8k / 16k. Underlies Figure~\ref{fig:context-length} (suite-level deltas) and Figure~\ref{fig:context-length-variance} (cross-seed std).}
    \label{tab:appendix-context-public}
\end{table}

\begin{table}[h]
    \centering
    \scriptsize
    \resizebox{\linewidth}{!}{
    \begin{tabular}{lcccccc}
        \toprule
        & \multicolumn{2}{c}{4k context} & \multicolumn{2}{c}{8k context} & \multicolumn{2}{c}{16k context} \\
        \cmidrule(lr){2-3} \cmidrule(lr){4-5} \cmidrule(lr){6-7}
        Capability & MAmmoTH-VL 2B & Datology 2B & MAmmoTH-VL 2B & Datology 2B & MAmmoTH-VL 2B & Datology 2B \\
        \midrule
        Chart        & 39.8 {\scriptsize $\pm$ 0.5}  & 51.7 {\scriptsize $\pm$ 0.9} & 37.8 {\scriptsize $\pm$ 0.7}  & 38.4 {\scriptsize $\pm$ 0.3} & 36.5 {\scriptsize $\pm$ 1.8}  & 39.9 {\scriptsize $\pm$ 0.5} \\
        Counting     & 80.0 {\scriptsize $\pm$ 1.4}  & 82.5 {\scriptsize $\pm$ 0.8} & 77.3 {\scriptsize $\pm$ 1.8}  & 82.3 {\scriptsize $\pm$ 0.4} & 76.7 {\scriptsize $\pm$ 0.5}  & 81.6 {\scriptsize $\pm$ 0.2} \\
        Document     & 39.5 {\scriptsize $\pm$ 0.9}  & 49.1 {\scriptsize $\pm$ 0.1} & 37.5 {\scriptsize $\pm$ 1.6}  & 41.4 {\scriptsize $\pm$ 0.5} & 37.9 {\scriptsize $\pm$ 0.7}  & 39.8 {\scriptsize $\pm$ 0.2} \\
        General      & 55.9 {\scriptsize $\pm$ 0.5}  & 57.5 {\scriptsize $\pm$ 0.1} & 55.5 {\scriptsize $\pm$ 0.3}  & 56.3 {\scriptsize $\pm$ 0.7} & 52.8 {\scriptsize $\pm$ 0.4}  & 54.5 {\scriptsize $\pm$ 0.9} \\
        Grounding    & 18.3 {\scriptsize $\pm$ 14.2} & 75.4 {\scriptsize $\pm$ 1.7} & 19.9 {\scriptsize $\pm$ 12.6} & 72.3 {\scriptsize $\pm$ 0.9} & 8.5 {\scriptsize $\pm$ 9.2}   & 70.4 {\scriptsize $\pm$ 0.4} \\
        Math         & 16.8 {\scriptsize $\pm$ 0.2}  & 19.1 {\scriptsize $\pm$ 0.3} & 16.5 {\scriptsize $\pm$ 0.8}  & 15.8 {\scriptsize $\pm$ 0.2} & 16.3 {\scriptsize $\pm$ 0.9}  & 17.1 {\scriptsize $\pm$ 0.9} \\
        Scene        & 61.1 {\scriptsize $\pm$ 0.2}  & 67.8 {\scriptsize $\pm$ 0.5} & 57.0 {\scriptsize $\pm$ 3.1}  & 62.1 {\scriptsize $\pm$ 0.5} & 55.5 {\scriptsize $\pm$ 2.0}  & 61.7 {\scriptsize $\pm$ 1.6} \\
        Spatial      & 41.0 {\scriptsize $\pm$ 2.6}  & 45.3 {\scriptsize $\pm$ 2.6} & 36.2 {\scriptsize $\pm$ 5.2}  & 42.7 {\scriptsize $\pm$ 2.0} & 31.4 {\scriptsize $\pm$ 7.4}  & 43.2 {\scriptsize $\pm$ 2.3} \\
        Table        & 36.2 {\scriptsize $\pm$ 1.8}  & 42.4 {\scriptsize $\pm$ 0.5} & 35.3 {\scriptsize $\pm$ 6.0}  & 38.9 {\scriptsize $\pm$ 0.5} & 33.4 {\scriptsize $\pm$ 0.9}  & 33.5 {\scriptsize $\pm$ 1.8} \\
        \midrule
        \textbf{Avg} & \textbf{43.2} {\scriptsize $\pm$ 1.7} & \textbf{54.5} {\scriptsize $\pm$ 0.5} & \textbf{41.4} {\scriptsize $\pm$ 2.3} & \textbf{50.0} {\scriptsize $\pm$ 0.5} & \textbf{38.8} {\scriptsize $\pm$ 1.8} & \textbf{49.1} {\scriptsize $\pm$ 0.6} \\
        \bottomrule
    \end{tabular}
    }
    \caption{Per-capability scores on the 9-capability \datbench{} suite at 2B across the 4k / 8k / 16k context-length sweep. Cells are mean $\pm$ std across training seeds (Datology has 3 / 2 / 2 seeds at 4k / 8k / 16k respectively). The Avg row gives a $+11.3$ / $+8.6$ / $+10.3$pp curation gain at 4k / 8k / 16k. Underlies Figure~\ref{fig:context-length} and Figure~\ref{fig:context-length-variance}.}
    \label{tab:appendix-context-datbench}
\end{table}

\subsection{Case-study and OOD raw numbers}
\label{app:results-case-studies}

This subsection lifts out the per-seed and per-metric tables for grounding (\S\ref{subsec:grounding-deep-dive}), counting, and BLINK (\S\ref{sec:main-results} OOD subsection), which sit in the appendix to keep the main text readable. Counting is included here in full (no main-text subsection); the grounding deep-dive in \S\ref{subsec:grounding-deep-dive} cross-references this appendix as a parallel instance of the same coherent-across-metrics pattern.

\subsubsection{Grounding (RefCOCO) per-metric}
\label{app:results-grounding}

\begin{table}[h]
    \centering
    \small
    \begin{tabular}{lcc}
        \toprule
        Metric & MAmmoTH-VL Mean $\pm$ Std & Datology Mean $\pm$ Std \\
        \midrule
        \texttt{recall@0.3} & $57.61 \pm 16.99$ & $85.92 \pm 0.68$ \\
        \texttt{recall@0.5} & $41.04 \pm 13.69$ & $80.29 \pm 1.01$ \\
        \texttt{recall@0.7} & $14.55 \pm 4.85$ & $69.96 \pm 0.88$ \\
        \texttt{center\_acc} & $69.72 \pm 12.25$ & $91.03 \pm 0.40$ \\
        \bottomrule
    \end{tabular}
    \caption{Grounding metrics on the RefCOCO subset, \base vs.\ \curated, in percentage units. Aggregated across the three RefCOCO splits used in \S\ref{subsec:grounding-deep-dive}.}
    \label{tab:grounding-metrics}
\end{table}

\subsubsection{Counting (CountBench) per-seed}
\label{app:results-counting}

CountBench scores at exact match, within-1, and within-3 tolerances; jointly they distinguish a narrow gain (one tolerance only) from a broad one. Curation lifts every tolerance: exact match from $86.08 \pm 0.51\%$ to $89.88 \pm 0.92\%$ ($+3.8$pp), within-1 from $95.32 \pm 0.89\%$ to $97.22 \pm 0.51\%$ ($+1.9$pp), and within-3 from $97.90 \pm 0.42\%$ to $99.12 \pm 0.12\%$ ($+1.2$pp). Every \curated seed exceeds every \base seed at all three tolerances. The gain shows up at the strictest threshold and carries through the looser ones, the same pattern observed in the grounding deep-dive (\S\ref{subsec:grounding-deep-dive}).

\begin{table}[h]
    \centering
    \small
    \begin{tabular}{lccc}
        \toprule
         & Exact & Within-1 & Within-3 \\
        \midrule
        MAmmoTH-VL ts0 & 85.54 & 95.93 & 98.37 \\
        MAmmoTH-VL ts1 & 86.15 & 94.30 & 97.76 \\
        MAmmoTH-VL ts2 & 86.56 & 95.72 & 97.56 \\
        MAmmoTH-VL (Mean $\pm$ Std) & $86.08 \pm 0.51$ & $95.32 \pm 0.89$ & $97.90 \pm 0.42$ \\
        \midrule
        Datology ts0 & 89.00 & 96.74 & 98.98 \\
        Datology ts1 & 89.82 & 97.15 & 99.19 \\
        Datology ts2 & 90.84 & 97.76 & 99.19 \\
        Datology (Mean $\pm$ Std) & $89.88 \pm 0.92$ & $97.22 \pm 0.51$ & $99.12 \pm 0.12$ \\
        \midrule
        Abs.\ Gain & $+3.80$ & $+1.90$ & $+1.22$ \\
        \bottomrule
    \end{tabular}
    \caption{CountBench per-seed results across exact, within-1, and within-3 tolerances; all values in percentage units.}
    \label{tab:counting-seeded}
\end{table}

\subsubsection{BLINK per-seed}
\label{app:results-blink}

\begin{table}[h]
    \centering
    \small
    \begin{tabular}{lrrrrrr}
        \toprule
        Training Mix & ts0 & ts1 & ts2 & Mean & Std & Abs.\ Gain vs MAmmoTH-VL \\
        \midrule
        MAmmoTH-VL & 41.29 & 41.35 & 42.35 & 41.66 & 0.59 & +0.00 \\
        Datology & 44.03 & 44.87 & 45.34 & 44.75 & 0.67 & +3.09 \\
        \bottomrule
    \end{tabular}
    \caption{BLINK per-seed accuracy in percentage units. Aggregate \datbench{}-style accuracy across all BLINK categories; the per-category breakdown referenced in \S\ref{sec:main-results} is summarized inline rather than tabulated here.}
    \label{tab:blink-seeded}
\end{table}

\clearpage
\section{Limitations and Future Work}
\label{app:limitations}

The results in this paper are at 1B, 2B, and 4B parameter scales, on a single base corpus (MAmmoTH-VL-12M single-image subset), one backbone family (Qwen3 LM with SigLIP2 vision encoder), and one training recipe. The robustness evidence in \S\ref{sec:robustness} indicates the qualitative picture carries. A direct demonstration across backbones and on additional open VLM corpora is future work. The pipeline does not extend to multi-image or interleaved image-text training. The BLINK transfer in \S\ref{sec:main-results} indicates curation applied to those data types would further improve performance.

\end{document}